\documentclass{article}

\usepackage{PRIMEarxiv}

\usepackage[utf8]{inputenc} 
\usepackage[T1]{fontenc}    
\usepackage{hyperref}       
\usepackage{url}            
\usepackage{booktabs}       
\usepackage{amsfonts}       
\usepackage{nicefrac}       
\usepackage{microtype}      
\usepackage{lipsum}
\usepackage{fancyhdr}       
\usepackage{graphicx}       
\graphicspath{{media/}}     

\usepackage{graphicx}
\usepackage{caption}
\usepackage[list=true,labelformat=simple]{subcaption}
\usepackage{xcolor}
\usepackage{amsmath}

\title{A Novel Center-based Deep Contrastive Metric Learning Method for the Detection of Polymicrogyria in Pediatric Brain MRI
}

\author{
  Lingfeng Zhang \\
  School of Electrical Engineering and Computer Science \\
  University of Ottawa \\
  Ottawa, Canada\\
  \texttt{lzhan278@uottawa.ca} \\
   \And
  Nishard Abdeen \\
  Department of Medical Imaging \\
  Children's Hospital of Eastern Ontario \\
  Department of Radiology \\
  University of Ottawa \\
  Ottawa, Canada\\
  \texttt{nishard.abdeen@cheo.on.ca} \\
  \And
  Jochen Lang \\
  School of Electrical Engineering and Computer Science \\
  University of Ottawa \\
  Ottawa, Canada\\
  \texttt{jlang@uottawa.ca} \\
}

\begin{document}
\maketitle

\begin{abstract}
Polymicrogyria (PMG) is a disorder of cortical organization mainly seen in children, which can be associated with seizures, developmental delay and motor weakness. PMG is typically diagnosed on  magnetic resonance imaging (MRI) but some cases can be challenging to detect even for experienced radiologists. In this study, we create an open pediatric MRI dataset (PPMR) with PMG and controls from the Children’s Hospital of Eastern Ontario (CHEO), Ottawa, Canada. The differences between PMG MRIs and control MRIs are subtle and the true distribution of the features of the disease is unknown. This makes automatic detection of cases of potential PMG in MRI difficult. We propose an anomaly detection method based on a novel center-based deep contrastive metric learning loss function (cDCM) which enables the automatic detection of cases of potential PMG. Additionally, based on our proposed loss function, we customize a deep learning model structure that integrates dilated convolution, squeeze-and-excitation blocks and feature fusion for our PPMR dataset. Despite working with a small and imbalanced dataset our method achieves $92.01\%$ recall at $55.04\%$ precision. This will facilitate a computer aided tool for radiologists to select potential PMG MRIs. 
To the best of our knowledge, this research is the first to apply machine learning techniques to identify PMG from MRI only.

Our code will be available at: \href{https://github.com/RichardChangCA/Deep-Contrastive-Metric-Learning-Method-to-Detect-Polymicrogyria-in-Pediatric-Brain-MRI}{https://github.com/RichardChangCA/Deep-Contrastive-Metric-Learning-Method-to-Detect-Polymicrogyria-in-Pediatric-Brain-MRI}.

Our pediatric MRI dataset will be available at: \href{https://www.kaggle.com/datasets/lingfengzhang/pediatric-polymicrogyria-mri-dataset}{\nolinkurl{https://www.kaggle.com/datasets/lingfengzhang/pediatric-polymicrogyria-mri-dataset}}.
\end{abstract}


\section{Introduction}\label{chapter:introduction}

\begin{figure}[htb]
    \centering 
\begin{minipage}[t]{.4\textwidth}
\begin{subfigure}{\textwidth}
  \includegraphics[width=\linewidth]{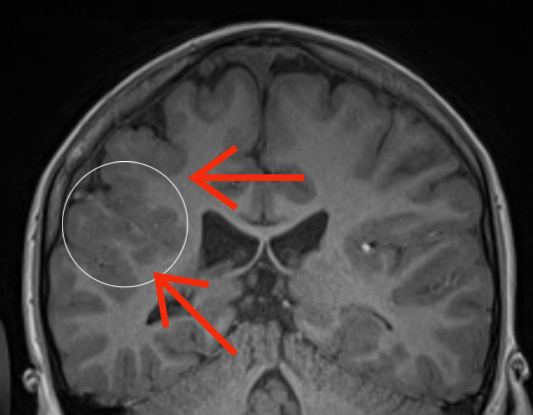}
  \caption{PMG MRI}
  \label{fig:polymicrogyria_mri}
\end{subfigure}
\end{minipage}\hfil
\begin{minipage}[t]{.4\textwidth}
\begin{subfigure}{\textwidth}
  \includegraphics[width=\linewidth]{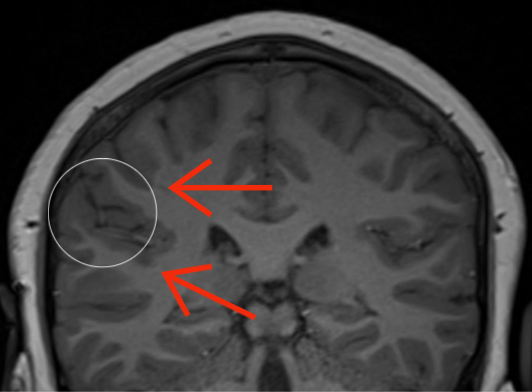}
  \caption{Control MRI}
  \label{fig:control_mri}
\end{subfigure}
\end{minipage}\hfil
\caption{Difference between PMG and normal brain on coronal T1 weighted MR images. The white circle on the first image shows thickened, irregular grey matter with numerous small gyri and shallow sulci, compared to normal grey matter in the white circle on the second image. The interface between cortical grey matter and white matter is irregular in PMG (red arrows on the first image) and smooth in normal brain (red arrows on the second image).}
\label{fig:polymicrogyria_control_demo}
\end{figure}

Polymicrogyria (PMG) is a cortical malformation characterized by irregular thickened grey matter in the cerebral cortex, with numerous small gyri, shallow sulci and loss of grey white matter differentiation 
(see Figure~\ref{fig:polymicrogyria_control_demo}).
Polymicrogyria is one of the most common malformations of cortical development and is most often related to genetic disorders although it can be caused by congenital viral infection~\cite{barkovich2010current}. PMG is a heterogeneous condition which can have variable distribution, extent, and severity~\cite{barkovich2010current}. 
Experienced neuroradiologists can accurately diagnose PMG using magnetic resonance imaging (MRI) but focal or more subtle PMG can be missed. Less experienced readers may not detect PMG with the same accuracy as it can be subtle, and computer aided diagnosis would be useful here.

Machine learning has been applied to cortical malformations in the pediatric brain, mainly in the context of fetal imaging~\cite{attallah2020deep} and epilepsy~\cite{sone2021clinical} and may 
contribute to computer-aided diagnosis.
There is however a lack of literature applying machine learning and in particular deep learning to the detection of polymicrogyria. 

The disease is rare in most clinical settings, while collecting normal pediatric brain MRI is relatively easy. We collected and annotated a small pediatric PMG MRI dataset (23 patients) from the Children's Hospital of Eastern Ontario (CHEO), Ottawa, Canada. 
This pediatric polymicrogyria MRI (PPMR) dataset is imbalanced because the majority of MRI are control 
and even for PMG samples the majority of MRI slices do not show evidence of PMG. We believe the composition of the PPMR dataset is a realistic reflection of the data likely to be encountered in clinical practice. In this paper, we introduce the PPMR dataset in detail. Because of the nature of PMG and the difficulty in collecting data, this dataset is unfortunately small and lacks the diversity of PMG. As it is difficult to evaluate testing results on such a dataset, we also introduce a modified CIFAR-10 dataset. This modified CIFAR-10 dataset mimics the challenges of our PPMR dataset, including the fact that the minority class samples in the training data are not representative and that the whole dataset is imbalanced.

Deep learning models trained with cross-entropy-based loss functions on small imbalanced datasets,
perform poorly when applied to real-word out-of-distribution samples.
Ano\-maly detection has the potential to mitigate this issue.  
This motivates the design of our novel cDCM loss function that combines anomaly detection and binary classification. 
Each input MRI is mapped to a latent representation where our loss function can "cluster" normal samples together, because normal images share similar features, and push anomaly samples further away from the "cluster".  
During inference, according to the idea of supervised contrastive learning~\cite{khosla2020supervised}, samples 
that have similar features as normal samples in the training data can be classified as normal; on the other hand, samples 
that have dissimilar features as normal samples in the training data can be classified as anomalies. 

Our novel cDCM loss function can partly mitigate the problem of generalization in deep learning, and our custom deep learning model structure can learn and select key features of MRI images automatically. In this paper, we compare our novel cDCM loss function to the Deep SAD loss function~\cite{ruff2019deep} and cross-entropy-based loss functions to demonstrate the benefits of our novel cDCM loss function. Our novel cDCM loss function can directly provide a good decision threshold that cannot be determined from training and validation data.  We also propose a custom deep learning model structure, combining dilated convolutions~\cite{yu2015multi}, squeeze-and-excitation~\cite{hu2018squeeze} blocks, and a multi-level feature fusion~\cite{huang2017densely} mechanism to be used with our novel cDCM loss function. 
We conduct an ablation study to confirm the significance of each component in the overall structure and show the utility and stability of our custom deep learning model structure, compared to some popular CNN backbones, such as EfficientNet and ResNet50.


In summary, the main contributions of our research are listed below:

\begin{itemize}
\item We propose a novel center-based deep contrastive metric learning loss function (cDCM) to 
work with a small and imbalanced dataset for which some testing data features do not appear in the training data.
\item Our method integrates the strengths of supervised binary classification and anomaly detection and to the best of our knowledge, we are the first to apply deep contrastive learning 
to PMG MRI classification. 
\item We design a custom convolutional neural network (CNN) structure which combines dilated convolutions~\cite{yu2015multi}, squeeze-and-excitation~\cite{hu2018squeeze} blocks, and a multi-level feature fusion~\cite{huang2017densely} mechanism. 
\item We make our
pediatric PMG MRI dataset (a.k.a. PPMR dataset) from the Children's Hospital of Eastern Ontario (CHEO), Ottawa, Canada
\end{itemize}

The remainder of this paper is organized as follows:
In Section~\ref{chapter:related_work}, we give a literature review, including machine learning in PMG images, classification of brain MRI, anomaly detection in medical imaging, and deep metric learning and deep contrastive learning. In Section~\ref{chapter:proposed_method}, we give a detailed illustration of our proposed method, including the novel cDCM loss function and custom model structure. In Section~\ref{chapter:experiments_and_results}, we introduce two datasets: a modified CIFAR-10 dataset and our PPMR dataset, and also describe training details. In Section~\ref{chapter:results_analysis_and_discussion}, we show the results of substantial experiments on these two datasets and also analyze and discuss the results of these experiments. Finally, we give a conclusion of our research and also consider some future works.

\section{Related Work}
\label{chapter:related_work}

\subsection{Machine Learning in Polymicrogyria Images}
\label{sec:machine_learning_in_polymicrogyria_images}

There is limited literature about applying machine learning methods on detecting pediatric PMG from MRIs~\cite{attallah2018detecting}. Attallah et al.~\cite{attallah2018detecting} proposed a machine learning pipeline for fetal brain MRI classification. This method can predict several brain abnormalities, including PMG, before the infant with fatal brain abnormalities is born. This machine learning pipeline consists of four steps: 
region of interest (ROI) segmentation, contrast enhancement, feature extraction, and classification by linear discriminate analysis (LDA), support vector machine (SVM), K-nearest neighbour (KNN) and ensemble classifiers.
Attallah et al.~\cite{attallah2020deep} 
proposed an improved method that combines CNN feature extraction based on transfer learning and conventional machine learning classifiers to detect fatal brain MRIs. 
We do not compare our method to Attallah et al.~\cite{attallah2020deep}, despite the fact that their method led to $88.6\%$ accuracy and $94\%$ AUC on their MRI, 
because transfer learning did not extract useful features from our PMG images in our tests. 
Hence, we could not apply the approach of Attallah et al.~\cite{attallah2020deep} as transfer learning is required.

Plonski et al.~\cite{plonski2017multi} collected 740 features, such as cortical thickness and surface area, from T1-weighted MRIs and then used feature selection methods and a logistic regression classifier to detect congenital brain malformations including PMG that lead to dyslexia. The best results are $66\%$ AUC and $65\%$ accuracy after 10-fold cross-validation. They demonstrated the main reason leading to these relatively poor results is that dyslexia is a heterogeneous syndrome and there is little consistency in univariate grey matter.

However, these methods above are not specific to the detection of PMG on MRIs, because they applied their methods to a dataset which only includes a small portion of PMG. In addition, these methods have not shown they have the ability to deal with the problem of imbalanced datasets and generalization to unseen features.

\subsection{Classification of Brain MRI}
\label{sec:classification_in_medical_imaging}

While many approaches for medical image classification exist, we focus our brief review on related deep learning methods.

Medical brain data is typically volumetric. Since 3D brain MRIs include more information compared to 2D brain MRIs because of the third dimension, some researchers applied deep learning models to 3D brain MRIs directly and achieved good results when they have sufficient 3D volumetric data~\cite{korolev2017residual,wegmayr2018classification}. Korolev et al.~\cite{korolev2017residual} proposed an end-to-end 3D CNN architecture to classify Alzheimer's disease from normal controls on the 3D Alzheimers Disease National Initiative (ADNI) dataset. Wegmayr et al. collected 3D brain MRIs from various sources to build a large 3D brain dataset and customized a 3D CNN architecture to classify several neurodegenerative diseases~\cite{wegmayr2018classification}.
However, in our PPMR dataset, we only have 23 PMG brains in total, which is not sufficient for training a 3D CNN model. We can regard each PMG brain as a series of PMG slices; then the number of data samples is sufficient to train a deep learning model.

In medical imaging, it is difficult to acquire a sufficient number of images because of a limited number of patients and because of privacy concerns~\cite{greenspan2016guest}. Deep learning models are easily overfitting to the training data if they are trained on relatively small datasets~\cite{pasupa2016comparison}. Consequently, deep learning models will lack the ability to generalize on unseen data. 
CNNs as feature extractors with 
conventional machine learning models can 
improve generalization~\cite{deepak2019brain}.
However, in our case, deep 
feature extractors 
and traditional image processing methods 
do not extract essential features from our PPMR data 
, because the differences between normal and PMG MRI are subtle. 

\subsection{Anomaly Detection in Medical Imaging}
\label{sec:anomaly_detection_in_medical_imaging}

Pathologies in medical images are often regarded as rare deviance (anomalies) because the majority of medical images are healthy (or normal) samples~\cite{grubbs1969procedures,tschuchnig2022anomaly}. 
Unlike binary classification, anomaly detection can deal with a relatively imbalanced dataset where normal samples are the majority and the diversity of anomaly samples is limited.


While semi-supervised and unsupervised deep learning methods are commonly used in anomaly 
detection~\cite{tschuchnig2022anomaly}
but may be limited for medical images which have complex or even indistinguishable features~\cite{gao2020handling}, such as PMG images.
Therefore, exploring anomaly-supervisory signals may help anomaly detection~\cite{pang2021deep}. Ruff et al. introduced a deep semi-supervised anomaly detection (Deep SAD)~\cite{ruff2019deep} which showed that adding a few labeled anomaly samples during training can achieve better results than the deep support vector data description (Deep SVDD)~\cite{ruff2018deep}, which is trained only on normal samples. Ding et al.~\cite{ding2022catching} proposed a novel open-set supervised anomaly detection method to detect seen anomalies and unseen anomalies. They 
created some pseudo anomaly samples which 
have similar features as real anomaly samples 
but are dissimilar to normal samples. However, in our PMG detection,
real anomaly samples are close to normal samples because the differences between PMG MRI and normal MRI are subtle. 
Hence, generating pseudo anomaly samples for our PMG detection may not be appropriate.


\subsection{Deep Metric Learning and Deep Contrastive Learning}
\label{sec:deep_metric_learning_and_deep_contrastive_learning}

%
Deep metric learning~\cite{kaya2019deep} and deep contrastive learning~\cite{khosla2020supervised} 
share similar ideas. Both of them learn embedding representations from inputs. However, deep metric learning was inspired by traditional metric learning methods~\cite{kulis2013metric}, i.e., t-SNE~\cite{van2008visualizing}, K-nearest neighbors~\cite{peterson2009k}, and relevant variants. Metric learning maps input data samples into embedding representations and then calculate distances between these embeddings for clustering or classification tasks. Deep metric learning replaces sophisticated manually designed mapping functions with deep neural networks. On the other hand, deep contrastive learning is a subset of deep metric learning. Deep contrastive learning methods learn embedding representations in a contrastive fashion~\cite{chopra2005learning}. It requires that similar data samples (data samples from the same class) should be close to each other in the embedding space. On the other hand, dissimilar data samples (data samples from different classes) should be far away from each other in the embedding space. 
Han et al.~\cite{han2021pneumonia} used a contrastive learning method to pre-train a feature extractor in a self-supervised way and added a classification head to fine-tune the whole model for pneumonia detection on chest X-ray images. Similarly, Chen et al.~\cite{chen2021momentum} also used a contrastive loss to train an encoder. But they adopted the pre-trained encoder for classification using a few-shot learning method to diagnose COVID-19 based on chest CT images. More similar works~\cite{shurrab2021self,sowrirajan2021moco,sriram2021covid,azizi2021big} used contrastive learning methods to learn better feature representations from medical images. Deep contrastive learning methods do not always calculate distance based on the learned feature embeddings. 
However, our method is a deep contrastive metric learning method.


\section{Proposed Method}
\label{chapter:proposed_method}


Cross-entropy-based loss functions have several limitations: they easily lead to overfitting during training on small datasets; they tend to predict testing samples as the majority class; and they lack generalization when encountering unseen features. 
To mitigate these issues, we propose a novel cDCM loss function for center-based deep contrastive metric learning and a custom convolutional neural network (CNN) structure for PMG detection in pediatric brain MRIs. Our novel cDCM loss function can improve model generalization, and the decision boundary of this loss function tends to produce high recall with acceptable precision over two datasets: the modified CIFAR-10 and the PPRM. Since the goal of our research is to design a computer-aided diagnosis method to identify possible PMG, and radiologists should focus on images predicted as possible PMG, recall is more important than precision in our setting.
More details about our novel cDCM loss function can be found in Section~\ref{sec:loss_function}. 


We customize a deep learning structure which is specifically designed for our PPMR dataset and compare our custom model with two popular CNN structures, in particular, to EfficientNet and to ResNet50. Since radiologists need to focus on both local features, such as gyri and boundary, and global features, such as spatial surroundings, our custom structure is designed to extract both local and global features. Moreover, our custom structure is composed of dilated convolutions~\cite{yu2015multi}, squeeze-and-excitation~\cite{hu2018squeeze} blocks and DenseNet-like~\cite{huang2017densely} feature fusion, to assist in extracting key features of PMG on MRIs. More details about our custom model structure are discussed in Section~\ref{sec:model_structure}.

\subsection{Loss Function}
\label{sec:loss_function}

Inspired by the loss function of Deep SAD~\cite{ruff2019deep} and the hinge loss~\cite{wu2007robust}, we propose a novel center-based deep contrastive metric learning loss function (cDCM), which has a clear pre-defined decision boundary to help make the final decision. Our deep learning model $f$ maps each input image to an n-dimensional latent representation.



\begin{figure}[htb]
    \centering
    \includegraphics[width=0.65\columnwidth]{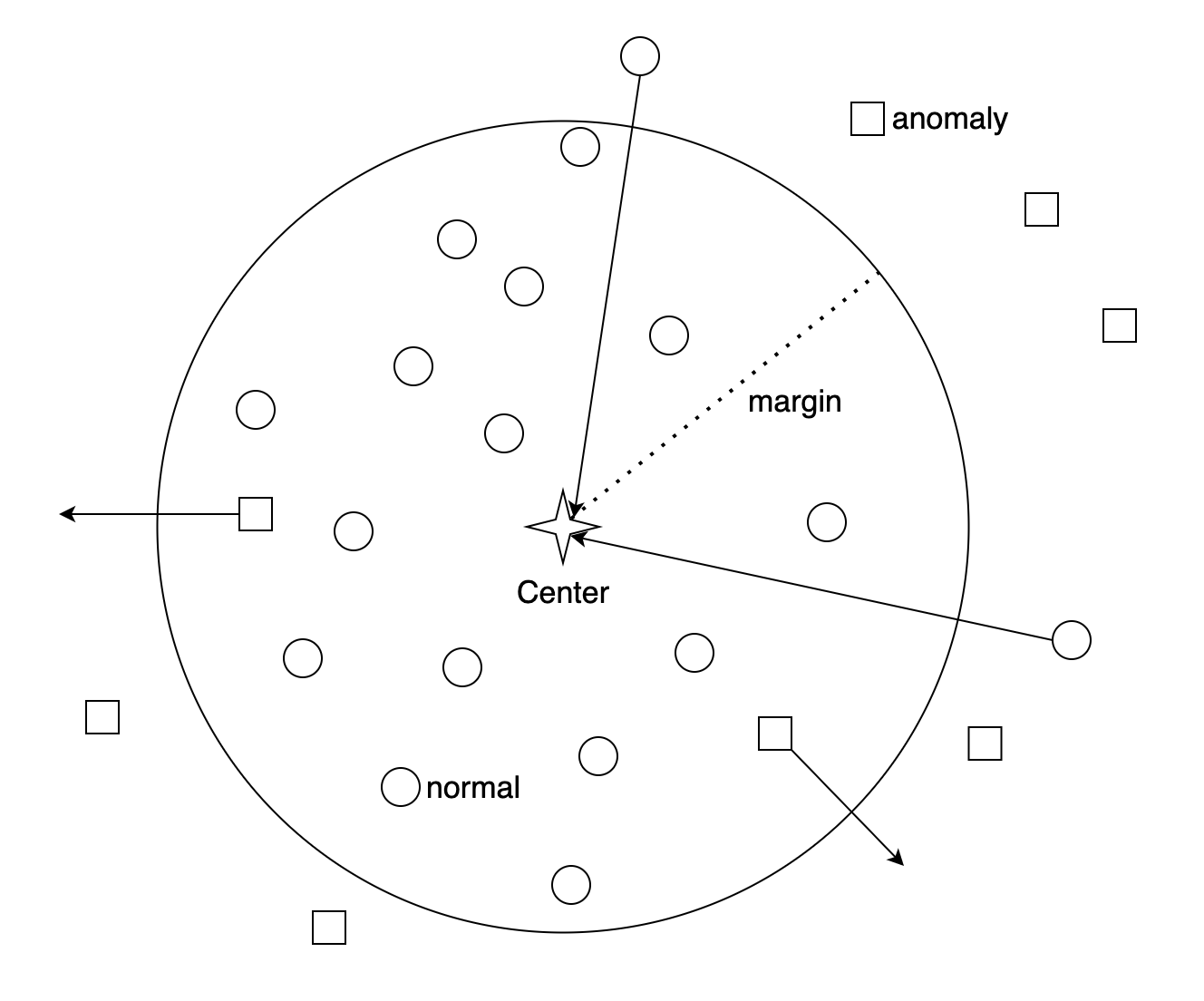}
    \caption{Illustration of our novel cDCM loss function, the four-pointed star represents the center $c$, anomaly samples in the latent representation are marked with a square, normal samples in the latent representation are represented by a small circle, and the large circle represents the decision boundary.}
    \label{fig:diagram}
\end{figure}

The basic idea of our novel cDCM loss function is shown in Figure~\ref{fig:diagram}. This loss can force negative (normal) samples to be close to the center $c$, and it can also push positive (anomaly) samples outside the circular margin centered at $c$.

We use the Euclidean norm to measure the distance between latent representations of samples and the center $c$, i.e.,
$
d_i = || l_i - c ||\ ,
$
where $l_i$ is the latent representation of sample i, and $d_i$ is the Euclidean norm between the latent representation and the center $c$.
%
%
To make sure the majority of positive (anomaly) samples are not close to the decision boundary (margin), we add the term $\frac{1}{1+e^{d_i-m}}$, which increases the gradients for positive (anomaly) samples which are outside of the circle but close to the decision boundary (margin). This term pushes positive (anomaly) samples further away from the center $c$. We also add the parameter $ \alpha $ to account for an imbalanced dataset. 
The loss function is then
\begin{equation}
Loss = \frac{1}{N}\sum\limits_{i=1}^{N}(1-y_i) * d_i + \frac{1}{M}\sum\limits_{j=1}^{M}\alpha * y_j * (max(0, m - d_j) +  \frac{1}{1+e^{d_j-m}})
\label{eqn:loss}
\end{equation}
where N is the number of normal samples in one batch; M is the number of anomaly samples in the same batch; m is the margin, which can also be regarded as the distance from the center $c$ to the decision boundary, and y is the label. Positive (anomaly) samples are labelled y=1, whereas y=0 is the label for negative (normal) samples.

\begin{figure}[htb]
    \centering
    \includegraphics[width=0.5\columnwidth]{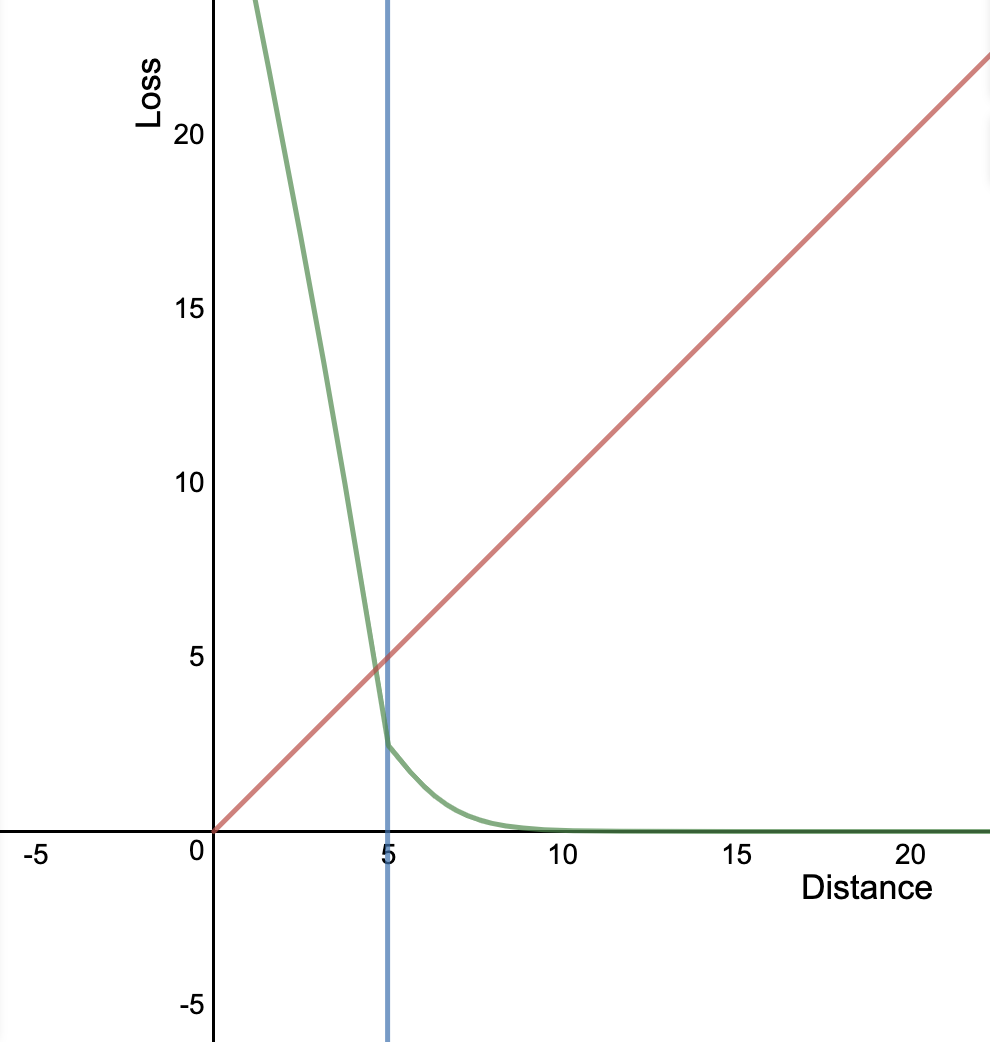}
    \caption{Loss value plot. The red line in the plot represents the loss of samples which are labeled as negative (normal), whereas the green line represents the loss of samples which are labeled as positive (anomaly). In addition, the blue line is the decision boundary (margin).}
    \label{fig:loss}
\end{figure}

The values of the different loss terms in Eqn.~\ref{eqn:loss} depending on the distance from the centre are plotted in Figure~\ref{fig:loss}.

A center $c$ is also employed in the deep contrastive metric learning methods Deep SVDD~\cite{ruff2018deep} and Deep SAD~\cite{ruff2019deep}. However, these methods use an autoencoder which is trained first and then, a center $c$ is calculated based on the average of latent representations of all normal samples in the training data. Deep SAD demonstrated fixing the center $c$ during training can achieve smoother and faster convergence~\cite{ruff2019deep}. In our research, we theoretically and experimentally prove (see Appendix~\ref{sec:DifferentCenter}) that in our loss the hyper-parameter center $c$ can be chosen randomly at initialization time, and then be fixed during training.


\subsection{Model Structure}
\label{sec:model_structure}

Although 
our cDCM loss function 
can address the generalization problem in small imbalanced datasets, it is still necessary to find a model which can extract essential features from our hospital PMG images. According to PMG disease characteristics in MRI,
gray matter contour in a patient's brain and gray matter location in relation to the whole brain are both important features to distinguish. 

\begin{figure}[htb]
    \centering
    \includegraphics[width=0.95\columnwidth]{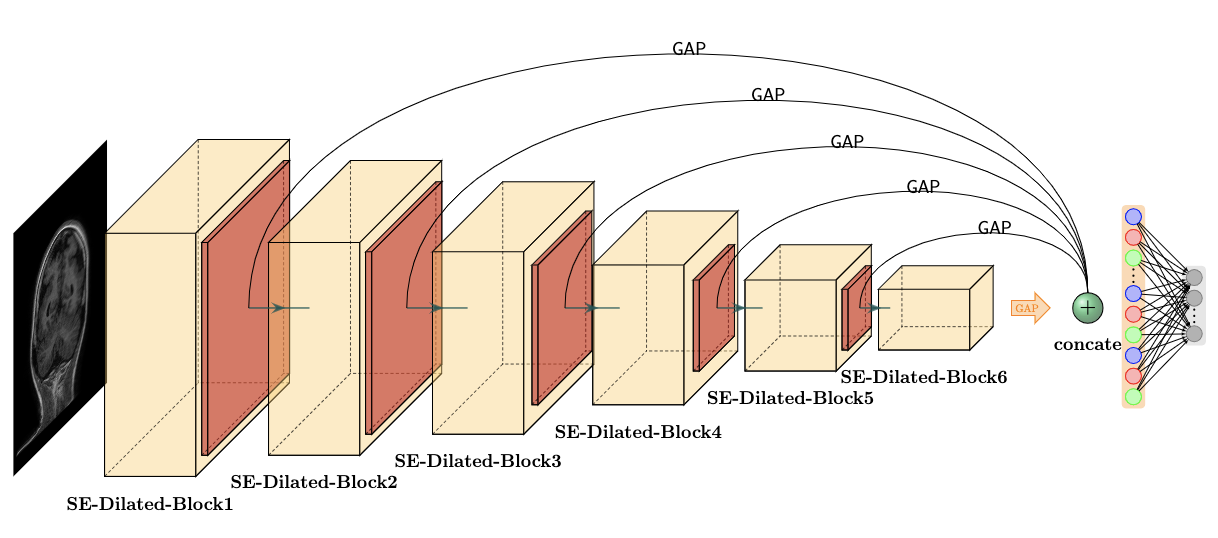}
    \caption{Model structure for PPMR dataset classification. GAP means global average pooling. Red rectangles represent max pooling.}
    \label{fig:whole_structure}
\end{figure}

\begin{figure}[htb]
    \centering
    \includegraphics[width=0.95\columnwidth]{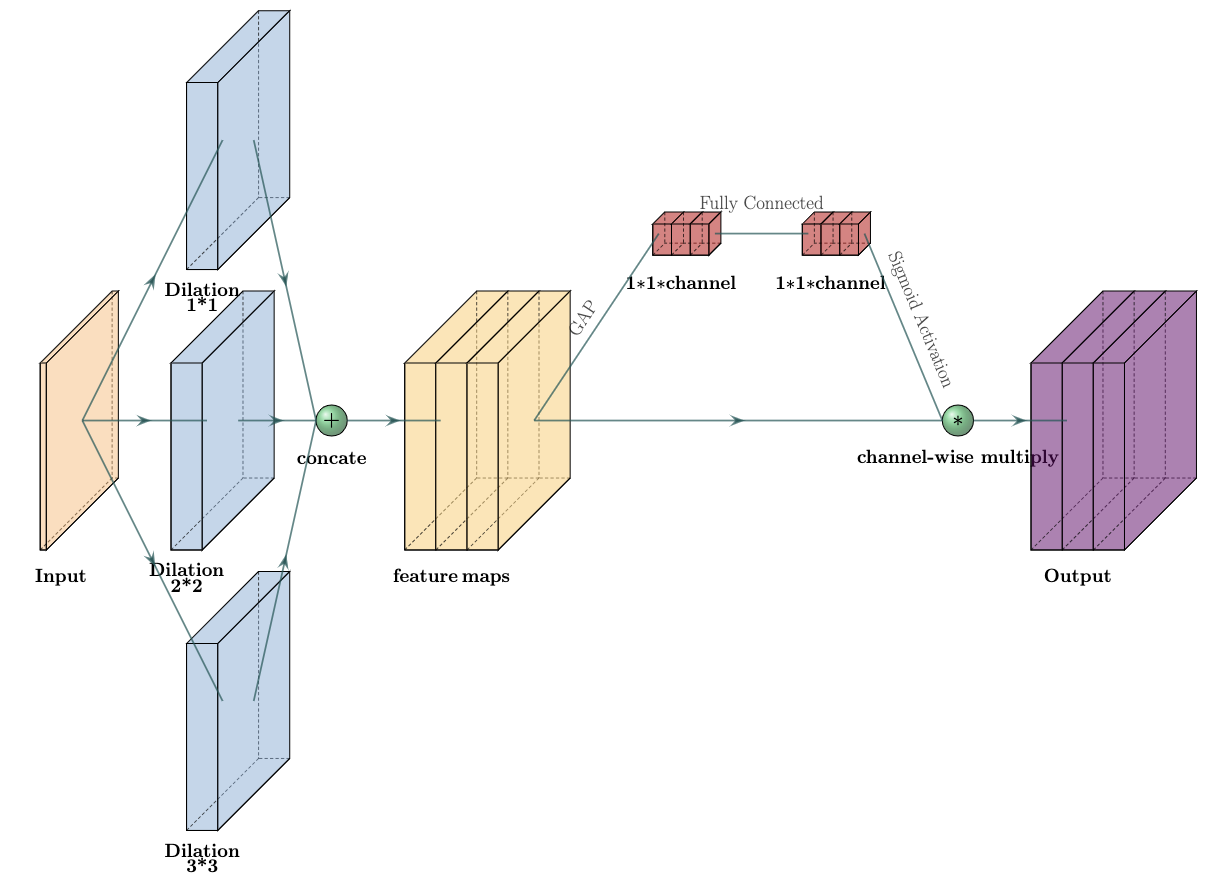}
    \caption{SE-Dilated-Block}
    \label{fig:SE_Dilated_Block}
\end{figure}

We design a custom CNN structure as shown in Figure~\ref{fig:whole_structure}. This model 
employs dilated convolution~\cite{yu2015multi}, squeeze-and-excitation (SE)~\cite{hu2018squeeze} and a channel-wise attention mechanism during feature fusion. In each SE-Dilated-Block (see Figure~\ref{fig:SE_Dilated_Block}), we use dilated convolution with three different dilation rates to extract features from different receptive fields. Dilated convolution with a small dilation rate 
can extract local features, the same as standard convolution, and dilated convolution with a large dilated rate can extract global features~\cite{hu2018squeeze}. 

However, although global and local features are both important, the importance of local features varies in different layers of the model. For example, shallow layers extract local features but deep layers extract more global features. To deal with this issue, we use the squeeze-and-excitation block, which performs like a channel-wise attention~\cite{chen2017sca}, to help us select important feature maps dynamically during training. In the whole structure, shallow SE-Dilated blocks extract local features, and deep SE-Dilated blocks extract global features. Fusing shallow features and deep features may improve overall performance. 
We use global average pooling to transform a series of 2D feature maps to 1D vectors for fusion via DenseNet-like~\cite{huang2017densely} connections. A Multi-Layer Perceptron (MLP) head maps features into a lower dimensional vector because we want to calculate an Euclidean distance metric in our novel cDCM loss function. Xia et al.~\cite{xia2015effectiveness} demonstrated Euclidean distance becomes less effective with increasing dimensionality. 
We also note that a low dimensional latent space helps to address the "Curse of Dimensionality" when calculating  distances~\cite{aggarwal2001surprising}. In addition, these fused features need to be selected for prediction. Hence, we use several MLP layers to simultaneously, transform features and reduce the dimensionality.


\section{Datasets and Experimental Settings}
\label{chapter:experiments_and_results}

In this section, we introduce two datasets: the pediatric polymicrogyria MRI (PPMR) dataset and the modified CIFAR10 dataset, clarify evaluation metrics, and give the experimental setting for these two datasets. 
%


\subsection{Pediatric Polymicrogyria MRI (PPMR) Dataset}

Patients under the age of 18 years with a diagnosis of PMG were identified by first searching the hospital radiology information system for polymicrogyria, then, the hospital picture archiving and communication system (PACS) was queried for MRI studies performed on these patients. The research protocol was approved by the Research Ethics Board of the Children’s Hospital of Eastern Ontario. Informed consent was waived given the retrospective nature of the study. 

The studies were reviewed by a fellowship trained pediatric neuroradiologist with 12 years of experience in reading pediatric brain MRI. If the presence of polymicrogyria was confirmed, the patient was included in the study. Images of the coronal 3D gradient echo T1 weighted sequence, which is the highest resolution sequence and best at showing cortical detail, were exported from the PACS system as JPEG images. One examination was chosen per patient. If there was more than one study, the most recent study was chosen. The imaging parameters for the coronal T1 weighted sequence are shown in Table~\ref{table:imaging_parameters_for_the_coronal_T1_weighted_sequence}.

\begin{table}
\caption{Imaging parameters for the coronal T1 weighted sequence}
\label{table:imaging_parameters_for_the_coronal_T1_weighted_sequence}
\begin{tabular}{lll}
\toprule
manufacturer & Siemens & General Electric \\
magnet & 3T Skyra & 1.5T Cigna\\
\midrule
repetition time (TR) & 2200 ms & 10.44 ms \\
inversion time (TI) & 1030 ms & 450 ms \\
echo time (TE) & 2.63 ms & 4.3 ms \\
matrix & 320*260 & 512*512 \\
slice thickness & 1.2 mm & 1.2 mm \\
FOV & 20*23 cm & 22*27 cm \\
\bottomrule
\end{tabular}
\end{table}

For each patient, three controls were chosen and matched for gender and age (within 6 months). As for the patients, the coronal 3D gradient echo T1 weighted sequence was chosen. The images anterior to the mid-coronal plane of the orbital globes and posterior to the torcula herophili were discarded as they did not adequately show grey and white matter. The patient images were again reviewed by the radiologist and labelled as to the presence or absence of PMG. 

This dataset contains 23 patients in total. 
Within each patients MRI, not all images show evidence of PMG and some slices are normal. Although the ratio between controls and patients is 3:1, the ratio between normal slices and anomaly slices is around 5:1. Each patient's brain includes around 150 scans on average. The numbers 
for each patient are shown in Appendix~\ref{appendix_2}.

We adapt a 5-fold double CV for the PPMR dataset and 
split these 23 patient brains into training, inner fold validation, and outer fold validation data, with a ratio of 15:4:4 at the patient level. 
We regard PMG scans in the inner fold validation data 
as 'seen anomaly'. 
PMG scans in outer fold validation data 
are considered 'unseen anomaly'. 

To avoid data leakage, 
all PMG scans of one certain patient must be only in training, inner fold validation, or outer fold validation data exclusively~\cite{yagis2019generalization,alinsaif20213d}. The patient-based data splitting method makes working with the small dataset more challenging but is crucial because of the similarity between neighboring slices for one patient.

\subsection{Modified CIFAR-10 Dataset}

We provide a modified CIFAR-10 dataset setting to mimic the challenges of the PPMR dataset. 
We use a modified CIFAR-10 dataset to perform exhaustive experiments, comparisons, and analysis. 

The CIFAR-10 dataset is a collection of color images with the resolution of $32 \times 32$ grouped into ten classes: airplane, automobile, bird, cat, deer, dog, frog, horse, ship and truck. Each class contains 5,000 training images and 1,000 testing images.
%
We pick one class as normal, and the rest of the 9 classes as anomalies, but only 5 anomaly classes appear in training and validation data. We call these classes that appear in training and validation data as {\em seen anomaly}. On the other hand, classes that only appear in testing data are called {\em unseen anomaly}.

Since our PPMR dataset is imbalanced and lacks data diversity, we modify the CIFAR-10 dataset to make it an imbalanced dataset. The imbalanced ratio in the modified CIFAR-10 dataset should be slightly different from the ratio between normal and anomalous samples in our PPMR dataset, which is 5:1. In our PPMR dataset, the diversity of anomalous samples is small because there are multiple consecutive MRIs which are similar to each other. 
In addition, the diversity of anomalous samples in the modified CIFAR-10 dataset ought to be larger than the anomalous diversity in the PPMR dataset if they have the same imbalanced ratio. 
Hence, we set the imbalance ratio as 10:1 of normal to anomalous samples for the modified CIFAR-10 dataset.

After modification, the amount of normal training data is $5,000*80\%=4,000$ and the amount of anomalous training samples is $4,000*10\%=400$. The amount of normal validation data is $5,000*20\%=1,000$ and the amount of anomaly validation data is $1,000*10\%=100$. In the testing data, we want to observe the distribution of prediction values on the whole, so we have not removed any data samples from testing data. Hence, the amount of normal testing data is $1,000$; and the amount of anomaly testing data is $1,000*9=9,000$.
One detailed example where the ship class with class ID 8 is regarded as the normal class is shown in Table~\ref{table:cifar_10_data_split}. In this case, the normal class has ID 8, seen anomaly classes have IDs 9,0,1,2 or 3 and unseen anomaly class have IDs 4,5,6 or 7.


\begin{table}
\caption{Data splitting example}
\label{table:cifar_10_data_split}
\begin{tabular}{llll}
\toprule
Data Splitting & Training & Validation & Testing \tabularnewline
\midrule
 Normal Data Amount & 4,000 & 1,000 & 1,000 \tabularnewline
 Anomaly Data Amount & 400 & 100 & 9,000 \tabularnewline
 Normal Class ID & 8 & 8 & 8 \tabularnewline
 Anomaly Class IDs & 9,0,1,2,3 & 9,0,1,2,3 & 9,0,1,2,3,4,5,6,7 \tabularnewline
\bottomrule
\end{tabular}
\end{table}

\subsection{Metric and Evaluation}

Accuracy, precision, recall, $F_\beta$ measure, area under the curve of ROC are commonly used to evaluate the performance of binary classification tasks. The recall, also known as sensitivity, or true positive rate, is regarded as the most important metric in medical image classification~\cite{hicks2022evaluation}. 
The goal for our model is to miss as few positive (anomaly) samples as possible. It is unreasonable to only take recall into account because it is trivial for a classifier to attain $100\%$ recall by simply predicting all samples as positive. Hence, the $F_2$ measure is chosen as our main evaluation metric.
We will also report other metrics such as accuracy, area under receiver operating characteristic (ROC) curve~\cite{mandrekar2010receiver}, etc. 


\subsection{Experimental Settings on the Modified CIFAR-10 Dataset}
\label{section:Experimental_Settings_CIFAR}

Our method uses an end-to-end training strategy. In our novel cDCM loss training, we employ the same  LeNet-type~\cite{lecun1989backpropagation} convolutional neural networks (CNN) to train our model on the modified CIFAR-10 dataset as Ruff et al.\ have used in their corresponding Deep SAD experiments. More hyper-parameters settings are shown in Table~\ref{table:hyper_parameters_setting_cifar10}.



\begin{table}
\caption{Hyper-parameters setting for the modified CIFAR-10 dataset}
\label{table:hyper_parameters_setting_cifar10}
\begin{tabular}{ll}
\toprule
Hyper-parameters & Value\tabularnewline
\midrule
margin & 5\tabularnewline
theta & 5\tabularnewline
Latent Representation Dimension & 128\tabularnewline
CNN Kernel Size & 5\tabularnewline
Leaky ReLU ratio & 0.01\tabularnewline
Optimizer & Adam\tabularnewline
Initial Learning Rate & 1e-3\tabularnewline
Batch Size & 128\tabularnewline
Maximum Training Epochs & 400\tabularnewline
\bottomrule
\end{tabular}
\end{table}

As for the cross-entropy-based loss training, we add one more fully-connected dense layer to transform the 128-dimensional vector into one probability logit value with the help of a sigmoid activation.

Since modified CIFAR-10 training and validation data lack diversity, training models is easy for a model to overfit to the training data. To relieve overfitting, we choose four different data augmentation methods (i.e., rotate, shift, shear, and zoom) during training. 
We minimize the loss, however when evaluating the corresponding model parameters on the validation set, we rely on the AUCROC metric. Therefore, we save the model parameters with the best metric on the validation data during training. 
After inference, we find we can get a high recall and acceptable precision. 

%

During training, we decay the learning rate with a ratio of 0.5 when validation loss is on a plateau (no decrease) within 900 iterations. 
We set a the number of training epochs to 400 and apply early stopping when the validation loss does not decrease within 4,500 iterations. 
%
To obtain more stable results, we train each model 5 times with random weight initialization and then average the results of these 5 experiments. We choose the Xavier uniform initializer (default initializer for convolution in TensorFlow). Models for the modified CIFAR-10 dataset are implemented in TensorFlow 2.8 and trained on Google Colab Pro+.

\subsection{Experimental Settings on Our Pediatric Polymicrogyria MRI (PPMR) Dataset}
\label{section:Experimental_Settings_PPMR}

The input image shape in our PPMR dataset is $256 \times 256 \times 3$, and these images are normalized to the range from 0 to 1.



Some work~\cite{shree2018identification,sathi2020hybrid} used skull stripping as one pre-processing step on the brain tumor challenges. But we do not choose this pre-processing step because it may lose some essential information of PMG, such as parts of gray matters close to the brain skull. 

Models on our PPMR dataset are implemented in TensorFlow 2.4 and trained on a single NVIDIA GeForce RTX 3090 24GB GPU. We use a batch size of 64 in the PPMR dataset rather than 128 as in the modified CIFAR-10 dataset to fit the model and data into memory. In addition, we choose all 1.0 as the center representation; CNN kernel size of 3, except for the dilated convolution; and a dropout ratio of 0.2. The remaining hyper-parameters settings are the same as in Table~\ref{table:hyper_parameters_setting_cifar10}. Training strategies, including data augmentation, learning rate decay on a plateau, early stopping, saving model weights with the best validation AUCROC, and so on, are all the same as for training on the modified CIFAR-10 dataset.

Different from averaging five randomly initialized training results as for the modified CIFAR-10 dataset, 
%
%
we utilize 5-fold double cross-validation in our PPMR dataset. 
We utilize four inner folds with different validation data for each outer fold. Since our whole PPMR dataset is relatively small, validation data may be biased or be not representative. In addition, the poor quality of validation data will affect model generalization. If we want to obtain a model with good generalization, we have to save model weights where validation performance is at least good enough. Therefore, in each outer fold inference phase, we pick the model with the best validation AUCROC value in its corresponding 4 inner folds.

\section{Experimental Results, Analysis and Discussion}
\label{chapter:results_analysis_and_discussion}

In this section, we show comprehensive experimental results. To explore the advantages of our cDCM loss function, we perform experiments to compare our cDCM loss function with cross-entropy-based loss functions and with the Deep SAD loss function on the modified CIFAR-10 dataset in Section~\ref{section:Comparison_with_Cross_Entropy_based_Loss_Functions} and in Section~\ref{section:Comparison_with_the_State_of_the_art_Deep_SAD_Loss_Function}, respectively. 
Our novel cDCM loss function is inspired by the Deep SAD loss function.
We show that our novel cDCM loss function surpasses the state of the art. 
We then apply our cDCM loss function to the PPMR dataset and apply our customized deep learning model to the PPMR dataset. In addition, in Section~\ref{section:Comparison_between_our_model_and_other_CNN_models}, we also 
compare our custom deep learning model with two popular CNN models, in particular, with EfficientNet and ResNet50, 
and in Section~\ref{sec:ablation_study}, we conduct an ablation study.

\subsection{Comparison with Cross-Entropy-based Loss Functions}
\label{section:Comparison_with_Cross_Entropy_based_Loss_Functions}

We compare the modified CIFAR-10 dataset performance of our cDCM loss and several cross-entropy-based loss functions, i.e., binary cross-entropy loss (BCE), weighted binary cross-entropy loss (WBCE), and focal loss~\cite{lin2017focal}.
The weighted binary cross entropy loss puts more weight on the minority class in the loss function.
Focal loss emphasizes learning from hard samples, including misclassified samples and minority class samples~\cite{lin2017focal}. 

\begin{figure}[htb]
    \centering
    \includegraphics[width=0.9\columnwidth]{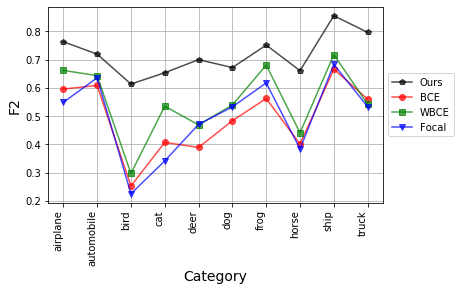}
    \caption{Comparison of $F_2$ measure results between our loss function and three cross-entropy-based loss functions on modified CIFAR-10 dataset at the pre-defined decision threholds.}
\label{fig:fmeasure_on_modified_cifar10}
\end{figure}

In the modified CIFAR-10 dataset, the distribution of the training and the testing data are not the same because of the unseen anomalies during training. 
Consequently, we choose the default threshold 0.5 for BCE, WBCE, and focal losses; and choose the hyper-parameter margin as the decision threshold for our cDCM loss.
Comparative results can be found in Table~\ref{table:comparison_with_cross_entropy_loss_on_modified_cifar10}. 
We can see AUCROC values of our novel cDCM loss and three cross-entropy-based loss functions are similar, but our cDCM loss function has an advantage at the predefined decision thresholds. In Figure~\ref{fig:fmeasure_on_modified_cifar10}, we can see that our cDCM loss achieves better $F_2$ measure values in all classes, compared with other loss functions. Our novel cDCM loss function tends to provide a higher recall and only slightly lower precision.

\begin{table*}
\caption{Comparison between our novel cDCM loss function, three cross-entropy-based loss functions, and the Deep SAD loss function on the modified CIFAR-10 dataset. Each result value in the table is the average of 5 experiments with different weight initialization.}
\label{table:comparison_with_cross_entropy_loss_on_modified_cifar10}
\begin{tabular}{lllllll}
  \toprule
Normal Class & Loss & Precision & Recall & $F_2$ measure & Accuracy & AUCROC \\
\midrule
airplane & Ours & 0.9841 $\pm0.0048$ & \textbf{0.7239} $\pm0.0594$ & \textbf{0.7637} $\pm0.0530$ & 0.7407 $\pm0.0496$ & 0.8929 $\pm0.0058$ \\
 airplane & BCE & 0.9929 $\pm0.0010$ & 0.5423 $\pm0.0297$ & 0.5962 $\pm0.0286$ & 0.5845 $\pm0.0263$ & 0.8870 $\pm0.0034$ \\
 airplane & WBCE & 0.9919 $\pm0.0031$ & 0.6127 $\pm0.0831$ & 0.6622 $\pm0.0772$ & 0.6467 $\pm0.0725$ & 0.8935 $\pm0.0100$ \\
 airplane & Focal & 0.9937 $\pm0.0011$ & 0.4956 $\pm0.0941$ & 0.5492 $\pm0.0931$ & 0.5431 $\pm0.0837$ & 0.8838 $\pm0.0049$ \\
 airplane & Deep SAD & 0.9836 $\pm0.0011$ & 0.7128 $\pm0.0236$ & 0.7543 $\pm0.0211$ & 0.7308 $\pm0.0203$ & 0.8840 $\pm0.0060$ \\
 \midrule
 automobile & Ours & 0.9934 $\pm0.0011$ & 0.6735 $\pm0.0172$ & 0.7198 $\pm0.0157$ & 0.7021 $\pm0.0150$ & 0.8927 $\pm0.0050$ \\
 automobile & BCE & 0.9978 $\pm0.0003$ & 0.5552 $\pm0.0262$ & 0.6091 $\pm0.0253$ & 0.5985 $\pm0.0234$ & 0.9162 $\pm0.0054$ \\
 automobile & WBCE & 0.9972 $\pm0.0012$ & 0.5915 $\pm0.0388$ & 0.6436 $\pm0.0364$ & 0.6308 $\pm0.0342$ & 0.9197 $\pm0.0037$ \\
 automobile & Focal & 0.9965 $\pm0.0012$ & 0.5832 $\pm0.0279$ & 0.6358 $\pm0.0265$ & 0.6230 $\pm0.0244$ & 0.9153 $\pm0.0027$ \\
 automobile & Deep SAD & 0.9876 $\pm0.0013$ & \textbf{0.7302} $\pm0.0159$ & \textbf{0.7703} $\pm0.0141$ & 0.7489 $\pm0.0138$ & 0.8744 $\pm0.0132$ \\
 \midrule
 bird & Ours & 0.9576 $\pm0.0091$ & \textbf{0.5641} $\pm0.0552$ & \textbf{0.6140} $\pm0.0518$ & 0.5849 $\pm0.0436$ & 0.7310 $\pm0.0190$ \\
 bird & BCE & 0.9827 $\pm0.0049$ & 0.2139 $\pm0.0465$ & 0.2531 $\pm0.0516$ & 0.2890 $\pm0.0405$ & 0.7363 $\pm0.0145$ \\
 bird & WBCE & 0.9815 $\pm0.0022$ & 0.2534 $\pm0.0204$ & 0.2975 $\pm0.0225$ & 0.3238 $\pm0.0179$ & 0.7483 $\pm0.0068$ \\
 bird & Focal & 0.9805 $\pm0.0052$ & 0.1896 $\pm0.0499$ & 0.2255 $\pm0.0562$ & 0.2671 $\pm0.0434$ & 0.7257 $\pm0.0107$ \\
 bird & Deep SAD & 0.9667 $\pm0.0013$ & 0.3246 $\pm0.0611$ & 0.3739 $\pm0.0646$ & 0.3823 $\pm0.0532$ & 0.7197 $\pm0.0111$ \\
 \midrule
 cat & Ours & 0.9662 $\pm0.0038$ & \textbf{0.6053} $\pm0.0479$ & \textbf{0.6538} $\pm0.0448$ & 0.6256 $\pm0.0396$ & 0.7741 $\pm0.0056$ \\
 cat & BCE & 0.9813 $\pm0.0057$ & 0.3566 $\pm0.0765$ & 0.4074 $\pm0.0811$ & 0.4145 $\pm0.0659$ & 0.7679 $\pm0.0096$ \\
 cat & WBCE & 0.9717 $\pm0.0155$ & 0.4881 $\pm0.1877$ & 0.5354 $\pm0.1787$ & 0.5243 $\pm0.1552$ & 0.7749 $\pm0.0186$ \\
 cat & Focal & 0.9825 $\pm0.0031$ & 0.2944 $\pm0.0427$ & 0.3419 $\pm0.0459$ & 0.3601 $\pm0.0370$ & 0.7548 $\pm0.0072$ \\
 cat & Deep SAD & 0.9750 $\pm0.0020$ & 0.4769 $\pm0.1082$ & 0.5290 $\pm0.1104$ & 0.5181 $\pm0.0944$ & 0.7609 $\pm0.0237$ \\
 \midrule
 deer & Ours & 0.9728 $\pm0.0018$ & \textbf{0.6546} $\pm0.0418$ & \textbf{0.7002} $\pm0.0385$ & 0.6727 $\pm0.0358$ & 0.8261 $\pm0.0058$ \\
 deer & BCE & 0.9927 $\pm0.0024$ & 0.3383 $\pm0.0255$ & 0.3896 $\pm0.0271$ & 0.4022 $\pm0.0227$ & 0.8251 $\pm0.0059$ \\
 deer & WBCE & 0.9873 $\pm0.0105$ & 0.4209 $\pm0.1828$ & 0.4692 $\pm0.1768$ & 0.4725 $\pm0.1561$ & 0.8267 $\pm0.0114$ \\
 deer & Focal & 0.9873 $\pm0.0126$ & 0.4224 $\pm0.1726$ & 0.4712 $\pm0.1681$ & 0.4736 $\pm0.1461$ & 0.8189 $\pm0.0094$ \\
 deer & Deep SAD & 0.9810 $\pm0.0028$ & 0.5420 $\pm0.0599$ & 0.5947 $\pm0.0578$ & 0.5782 $\pm0.0515$ & 0.8272 $\pm0.0072$ \\
 \midrule
 dog & Ours & 0.9852 $\pm0.0033$ & \textbf{0.6231} $\pm0.0402$ & \textbf{0.6722} $\pm0.0372$ & 0.6522 $\pm0.0337$ & 0.8624 $\pm0.0043$ \\
 dog & BCE & 0.9936 $\pm0.0026$ & 0.4293 $\pm0.0412$ & 0.4839 $\pm0.0421$ & 0.4838 $\pm0.0359$ & 0.8387 $\pm0.0054$ \\
 dog & WBCE & 0.9902 $\pm0.0051$ & 0.4856 $\pm0.0734$ & 0.5397 $\pm0.0715$ & 0.5324 $\pm0.0629$ & 0.8439 $\pm0.0058$ \\
 dog & Focal & 0.9885 $\pm0.0100$ & 0.4815 $\pm0.1358$ & 0.5331 $\pm0.1301$ & 0.5273 $\pm0.1151$ & 0.8389 $\pm0.0029$ \\
 dog & Deep SAD & 0.9849 $\pm0.0013$ & 0.6172 $\pm0.0139$ & 0.6670 $\pm0.0129$ & 0.6469 $\pm0.0118$ & 0.8454 $\pm0.0097$ \\
 \midrule
 frog & Ours & 0.9909 $\pm0.0016$ & \textbf{0.7090} $\pm0.0142$ & \textbf{0.7518} $\pm0.0126$ & 0.7323 $\pm0.0117$ & 0.9107 $\pm0.0056$ \\
 frog & BCE & 0.9985 $\pm0.0011$ & 0.5077 $\pm0.0357$ & 0.5628 $\pm0.0349$ & 0.5561 $\pm0.0316$ & 0.9066 $\pm0.0067$ \\
 frog & WBCE & 0.9943 $\pm0.0043$ & 0.6329 $\pm0.0969$ & 0.6810 $\pm0.0896$ & 0.6880 $\pm0.0924$ & 0.9139 $\pm0.0042$ \\
 frog & Focal & 0.9968 $\pm0.0020$ & 0.5651 $\pm0.0608$ & 0.6180 $\pm0.0587$ & 0.6069 $\pm0.0537$ & 0.9154 $\pm0.0052$ \\
 frog & Deep SAD & 0.9904 $\pm0.0013$ & 0.7047 $\pm0.0180$ & 0.7478 $\pm0.0163$ & 0.7281 $\pm0.0158$ & 0.9027 $\pm0.0113$ \\
 \midrule
 horse & Ours & 0.9847 $\pm0.0058$ & \textbf{0.6120} $\pm0.0689$ & \textbf{0.6613} $\pm0.0647$ & 0.6420 $\pm0.0583$ & 0.8510 $\pm0.0172$ \\
 horse & BCE & 0.9982 $\pm0.0008$ & 0.3493 $\pm0.0399$ & 0.4012 $\pm0.0419$ & 0.4138 $\pm0.0356$ & 0.8807 $\pm0.0075$ \\
 horse & WBCE & 0.9972 $\pm0.0016$ & 0.3883 $\pm0.0712$ & 0.4414 $\pm0.0729$ & 0.4484 $\pm0.0633$ & 0.8852 $\pm0.0029$ \\
 horse & Focal & 0.9981 $\pm0.0004$ & 0.3332 $\pm0.0186$ & 0.3844 $\pm0.0198$ & 0.3993 $\pm0.0167$ & 0.8854 $\pm0.0050$ \\
 horse & Deep SAD & 0.9844 $\pm0.0021$ & 0.5916 $\pm0.0388$ & 0.6427 $\pm0.0367$ & 0.6240 $\pm0.0338$ & 0.8236 $\pm0.0193$ \\
 \midrule
 ship & Ours & 0.9880 $\pm0.0028$ & 0.8281 $\pm0.0331$ & 0.8557 $\pm0.0280$ & 0.8362 $\pm0.0279$ & 0.9412 $\pm0.0062$ \\
 ship & BCE & 0.9954 $\pm0.0014$ & 0.6166 $\pm0.0733$ & 0.6665 $\pm0.0698$ & 0.6523 $\pm0.0650$ & 0.9373 $\pm0.0039$ \\
 ship & WBCE & 0.9940 $\pm0.0009$ & 0.6714 $\pm0.0179$ & 0.7179 $\pm0.0163$ & 0.7006 $\pm0.0155$ & 0.9395 $\pm0.0023$ \\
 ship & Focal & 0.9938 $\pm0.0015$ & 0.6347 $\pm0.0778$ & 0.6831 $\pm0.0730$ & 0.6675 $\pm0.0688$ & 0.9339 $\pm0.0023$ \\
 ship & Deep SAD & 0.9835 $\pm0.0023$ & \textbf{0.8411} $\pm0.0231$ & \textbf{0.8661} $\pm0.0194$ & 0.8443 $\pm0.0195$ & 0.9265 $\pm0.0092$ \\
 \midrule
 truck & Ours & 0.9828 $\pm0.0020$ & \textbf{0.7610} $\pm0.0298$ & \textbf{0.7968} $\pm0.0260$ & 0.7729 $\pm0.0250$ & 0.8974 $\pm0.0076$ \\
 truck & BCE & 0.9955 $\pm0.0009$ & 0.5057 $\pm0.0494$ & 0.5604 $\pm0.0489$ & 0.5530 $\pm0.0439$ & 0.9006 $\pm0.0073$ \\
 truck & WBCE & 0.9963 $\pm0.0008$ & 0.4874 $\pm0.0378$ & 0.5426 $\pm0.0375$ & 0.5370 $\pm0.0336$ & 0.9045 $\pm0.0030$ \\
 truck & Focal & 0.9954 $\pm0.0025$ & 0.4793 $\pm0.0764$ & 0.5337 $\pm0.0756$ & 0.5292 $\pm0.0673$ & 0.9032 $\pm0.0065$ \\
 truck & Deep SAD & 0.9824 $\pm0.0034$ & 0.7521 $\pm0.0154$ & 0.7891 $\pm0.0133$ & 0.7648 $\pm0.0122$ & 0.8769 $\pm0.0092$ \\
\bottomrule
\end{tabular}
\end{table*}

\begin{figure*}[htb]
    \centering 
\begin{minipage}[t]{.4\textwidth}
\begin{subfigure}{\textwidth}
  \includegraphics[width=\linewidth]{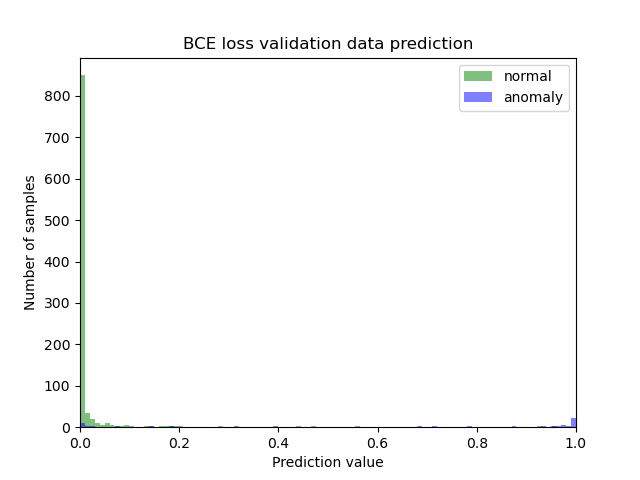}
  \caption{Validation data histogram with BCE loss}
  \label{fig:prediction_distribution_bce_val}
\end{subfigure}\hfil 
\begin{subfigure}{\textwidth}
  \includegraphics[width=\linewidth]{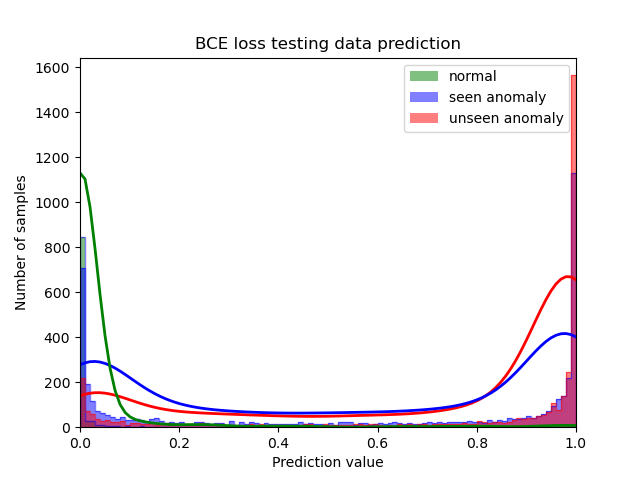}
  \caption{Testing data histogram with BCE loss}
  \label{fig:prediction_distribution_bce_test}
\end{subfigure}
\end{minipage}\hfil
\begin{minipage}[t]{.4\textwidth}
\begin{subfigure}{\textwidth}
  \includegraphics[width=\linewidth]{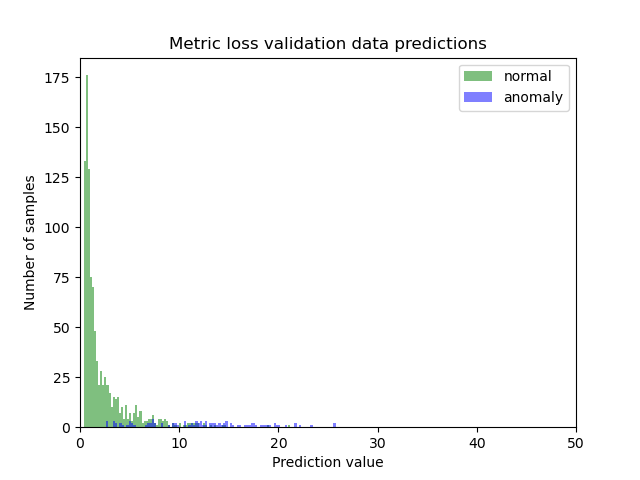}
  \caption{Validation data histogram with our novel cDCM loss}
  \label{fig:prediction_distribution_metric_val}
\end{subfigure}\hfil 
\begin{subfigure}{\textwidth}
  \includegraphics[width=\linewidth]{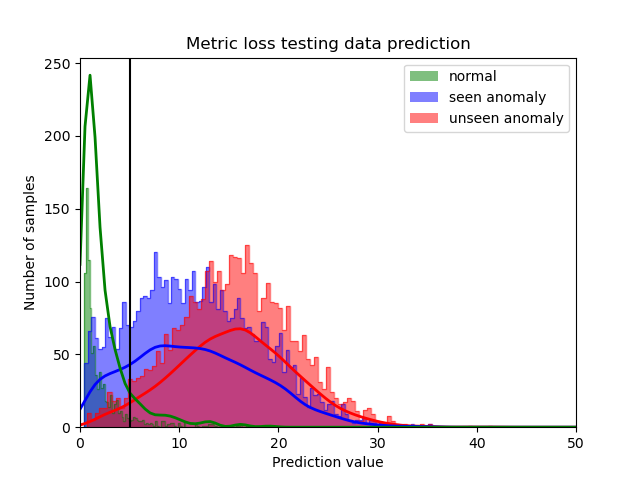}
  \caption{Testing data histogram with our novel cDCM loss}
  \label{fig:prediction_distribution_metric_test}
\end{subfigure}
\end{minipage}
\caption{Prediction histograms on the modified CIFAR-10 data samples with BCE loss and our novel cDCM loss. Here, we use the class 8 (ship) as normal.}
\label{fig:prediction_distribution}
\end{figure*}

We analyze prediction distributions on modified CIFAR-10 training, validation, and testing data separately. Results are shown in Figure~\ref{fig:prediction_distribution}. Figure~\ref{fig:prediction_distribution_bce_test} presents the testing data prediction distribution based on the BCE loss. Although more than half of anomaly samples are predicted correctly as anomalous, a large number of anomaly samples are predicted as normal. We can see that this BCE loss model seems to overfit the data; because the predictions on the validation data are perfect (see 
Figure~\ref{fig:prediction_distribution_bce_val}), but the prediction on the testing data is poor. 
However, using the resulting model obtained with our novel cDCM loss function results in a distribution of testing data predictions in the approximate shape of a normal distribution (see Figure~\ref{fig:prediction_distribution_metric_test}). Moreover, based on Figure~\ref{fig:prediction_distribution_bce_test}, simply moving the decision threshold for the BCE loss cannot trade-off precision and recall easily because the model trained on the BCE loss fails to recognize too many anomalous samples with high confidence. For example, if one prefers recall to precision, it is reasonable to choose a low threshold $\ll 0.5$ based on the Figure~\ref{fig:prediction_distribution_bce_val}. However, it does not improve recall too much on the testing data.


Figure~\ref{fig:prediction_distribution_metric_test} represents the testing data prediction distribution based on our novel cDCM loss model. The black vertical line at a value of 5 on the horizontal axis represents the margin (or decision boundary). We can see only a small part of the samples are predicted incorrectly. Moreover, the majority of normal samples are predicted close to the center $c$, and the majority of anomaly samples are predicted outside of the decision boundary. In our PPMR dataset, we do not have enough testing data to represent the whole distribution of anomaly samples, and our PPMR testing data are sampled from the whole distribution of anomaly samples. Based on Figure~\ref{fig:prediction_distribution_metric_test}, if testing data are randomly sampled from the whole distribution, recall remains still high, but precision will decrease dramatically. This is the main reason why our PPMR dataset tends to result in high recall and high AUCROC, but relatively low precision results when we directly use the predefined margin as the decision threshold. However, precision and recall can be simply balanced by moving the threshold with our novel cDCM loss function, although we prefer recall over precision in our application. 

\subsection{Comparison with Deep SAD}
\label{section:Comparison_with_the_State_of_the_art_Deep_SAD_Loss_Function}

Both the Deep SAD loss and our novel cDCM loss are center-based loss functions. However, the decision threshold for the Deep SAD loss is not defined. Zhou et al.~\cite{zhou2021vae} use $95\%$ percentile of distance values from all training normal samples to the center as the decision threshold. For this reason, we also choose this method to calculate the decision threshold for Deep SAD.

Comparisons of the modified CIFAR-10 dataset are also listed in Table ~\ref{table:comparison_with_cross_entropy_loss_on_modified_cifar10}. We can see recall values, $F_2$ measure values, accuracy values, and AUCROC values of our novel cDCM loss are better than the performance of Deep SAD loss in the majority of classes, although there is a small precision decrease. These results show the advantages of our novel cDCM loss function, compared to the Deep SAD loss function. This is probably because our cDCM loss function has a clear predefined decision threshold which can distinguish confusing samples but the Deep SAD loss function may not disentangle these confusing samples.

\begin{figure}[htb]
    \centering
    \includegraphics[width=0.95\columnwidth]{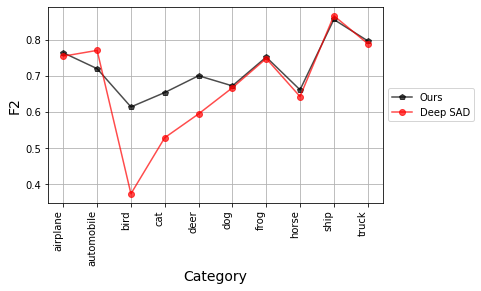}
    \caption{Comparison of $F_2$ measure results between our loss function and Deep SAD loss on modified CIFAR-10 dataset}
    \label{fig:fmeasure_compare_deepsad}
\end{figure}

\begin{figure}[htb]
    \centering
    \includegraphics[width=0.95\columnwidth]{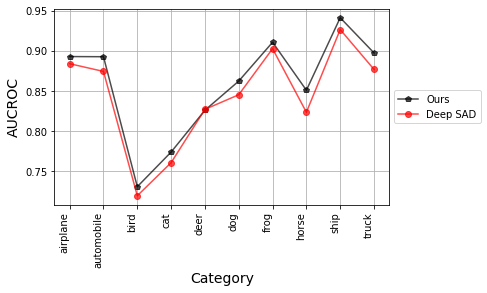}
    \caption{Comparison of AUCROC results between our loss function and Deep SAD loss on modified CIFAR-10 dataset}
    \label{fig:aucroc_compare_deepsad}
\end{figure}

Figure~\ref{fig:fmeasure_compare_deepsad} shows the comparison of $F_2$ measure results. In the majority of categories, our novel cDCM loss function performs better or slightly better than the Deep SAD loss, except for the category of automobile and ship. In addition, we can see there is a huge performance improvement when the category bird, cat, or deer is the normal class making our model performance more equal between classes. Hence, our novel cDCM loss function is more stable than the Deep SAD loss. Figure~\ref{fig:aucroc_compare_deepsad} shows the comparison of AUCROC results. We can see the AUCROC values of our novel cDCM loss function are higher than the AUCROC values of the Deep SAD loss function in all categories.

\begin{figure*}[htb]
    \centering 
\begin{minipage}[t]{.4\textwidth}
\begin{subfigure}{\textwidth}
  \includegraphics[width=\linewidth]{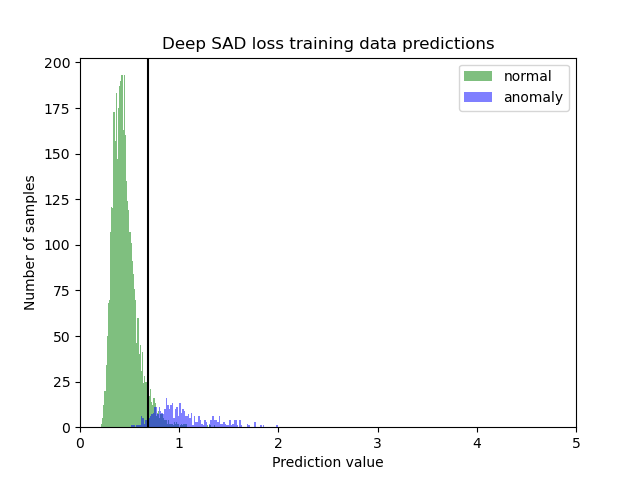}
  \caption{Training data histogram with Deep SAD loss}
  \label{fig:prediction_distribution_deepsad_train_2}
\end{subfigure}\hfil 
\begin{subfigure}{\textwidth}
  \includegraphics[width=\linewidth]{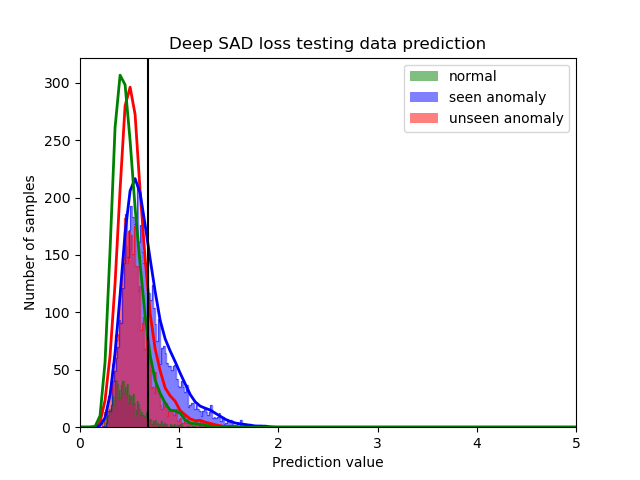}
  \caption{Testing data histogram with Deep SAD loss}
  \label{fig:prediction_distribution_deepsad_test_2_zoom}
\end{subfigure}
\end{minipage}\hfil
\begin{minipage}[t]{.4\textwidth}
\begin{subfigure}{\textwidth}
  \includegraphics[width=\linewidth]{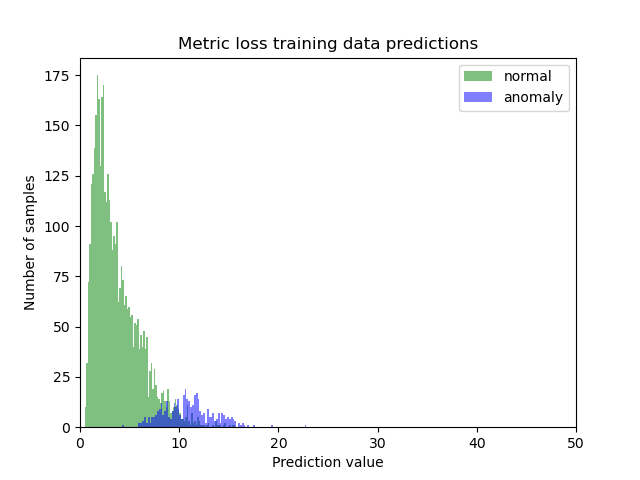}
  \caption{Training data histogram with our novel cDCM loss}
  \label{fig:prediction_distribution_metric_train_2}
\end{subfigure}\hfil 
\begin{subfigure}{\textwidth}
  \includegraphics[width=\linewidth]{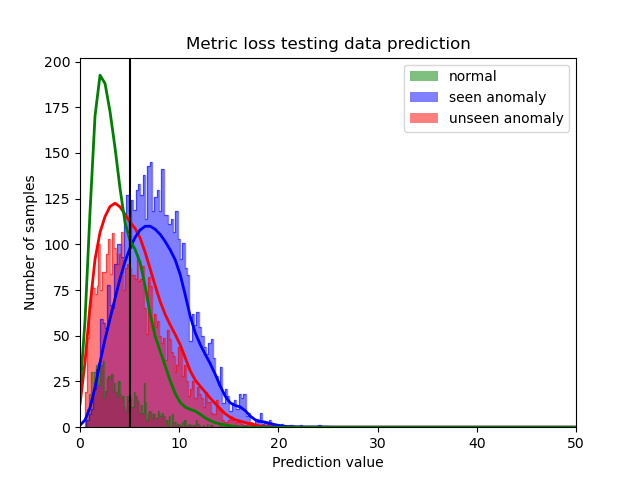}
  \caption{Testing data histogram with our novel cDCM loss}
  \label{fig:prediction_distribution_metric_test_2_again}
\end{subfigure}
\end{minipage}
\caption{Modified CIFAR-10 data samples prediction distribution with the Deep SAD loss and with our novel cDCM loss. The class 2 (bird) is the normal class. Please note: the x-axis is range from 0 to 5 in the left three plots, while the x-axis is range from 0 to 50 in the right three plots. The black vertical line is the decision threshold.}
\label{fig:prediction_distribution_deepsad_comparison_zoom}
\end{figure*}

In Table~\ref{table:comparison_with_cross_entropy_loss_on_modified_cifar10} and Figure~\ref{fig:fmeasure_compare_deepsad}, we can see when the class 2 (bird) is the normal class, the AUCROC scores of our loss function and Deep SAD loss function are similar (0.731 vs. 0.7197) but the $F_2$ measure values of these two loss functions have a large difference (0.6140 vs. 0.3739). To analyze this phenomenon, prediction distribution plots are shown in Figure~\ref{fig:prediction_distribution_deepsad_comparison_zoom}. In Figure~\ref{fig:prediction_distribution_deepsad_test_2_zoom}, the black vertical line is the decision threshold, which is the $95\%$ percentile of distance values from all training normal samples to the center. We can see distributions of normal, seen anomaly, and unseen anomaly are close to each other and the majority of values of these three distributions are all on the left side of the decision threshold. Moreover, distributions of normal and unseen anomaly are almost the same, which is not beneficial in distinguishing them. Different from the distributions with the Deep SAD loss, the distributions with our novel cDCM loss are wide and the majority of values of the seen anomaly sample distribution are on the right of the decision boundary, while the majority of values of the normal sample distribution are on the left side of the decision boundary.

Overall, there is probably no good decision threshold for the Deep SAD loss function, but we can define a decision threshold for our novel cDCM loss function first, and the deep learning model will optimize our novel cDCM loss function and separate normal and anomaly distributions into the two sides of the decision threshold. 
In our method a good decision threshold can be chosen for imbalanced datasets where the distribution of the anomaly samples is only partially known. This is the main reason why our novel cDCM loss can achieve better $F_2$ measure than the Deep SAD loss, although they have similar AUCROC scores.

Moreover, we perform experiments on the modified CIFAR-10 dataset in Appendix~\ref{sec:DifferentCentersComparison} to verify that choosing different centers in our novel cDCM loss function does not affect the final performance too much. Additionally, choosing different margins in our novel cDCM loss function also does not affect the final performance too much and these results are given in Appendix~\ref{sec:DifferentMarginsComparison}. In addition, we conduct the tests of statistical significance between 5 different loss functions on the modified CIFAR-10 dataset in Appendix~\ref{sec:tests_of_statistical_significance}.

\subsection{Main Results on the PPMR Dataset}
\label{section:Comparison_between_our_model_and_other_CNN_models}

We apply our cDCM loss in the training of all models on the PPMR dataset (except as specifically noted). The above experiments show that our novel cDCM loss function is able to predefine a better decision threshold than binary cross-entropy-based loss functions and the earlier Deep SAD loss on the modified CIFAR-10 dataset. 
Since our PPMR dataset is small, 
we have experimented with transfer learning 
including with EfficientNet and ResNet50 pretrained on ImageNet.
%
%
We find 
these models initialized with transfer learning will predict all samples as positive (anomaly).
Hence, we do not use transfer learning from the ImageNet dataset for our specific challenge.

As discussed in Section~\ref{sec:model_structure}, our custom model structure combines dilated convolution, squeeze-and-excitation block and feature fusion. We compare our model structure with some popular light-weight CNN backbones, i.e., EfficientNet~\cite{tan2019efficientnet} and ResNet50~\cite{he2016identity}. Since our PPMR dataset is relatively small and the model may easily overfit the data, light-weight model structures with a limited number of parameters are more reasonable for comparison. 

Results are shown in Table~\ref{table:model_comparison}. We can see all metrics, except for recall and $F_2$ measure, of our custom model structure are similar to but slightly worse than metrics of ResNet50. As for ResNet50, moving the threshold can also improve recall with decreasing precision. However, it is difficult to find an expected threshold without knowing the testing data in practice. Based on our pre-defined decision boundary, our model structure can directly achieve a high recall and an acceptable precision, and hence a good $F_2$ measure. Moreover, we can see standard deviations of all metrics except for accuracy, of our custom model structure are smaller than for the other two models; hence our custom model structure can achieve more stable results. More importantly, the final results of ResNet50 are unstable, not only because the standard deviations are relatively large, but also because we find that a NAN loss often occurs during training with our novel cDCM loss function. In these cases, one solution is training the model from scratch again. When we train the ResNet50 backbone based on our novel cDCM loss function, we find that, sometimes, this model predicts anomaly samples at an increasingly large distance to the center until the model predicts at least one normal sample as the anomaly to get NAN. 
We do not find any NAN loss issues during training our custom model or training EfficientNet, probably because convolutional layers 
are not as "deep". In our application, we prefer results with our custom model structure, mainly because results are stable and the $F_2$ measure of our custom model is higher than with the other two backbone models. There is however plenty of room to explore more advanced model structures based on our custom model structure in the future.

\begin{table*}[hbt]
\caption{Model comparison: average of 5-fold CV results with standard deviation.}
  \label{table:model_comparison}
  \begin{tabular}{llllll}
    \toprule
    Model Blocks &Precision &Recall &$F_2$ measure &Accuracy &AUCROC \\
    \midrule
    EfficientNet + cDCM Loss &0.5260 $\pm0.3243$ &0.6599 $\pm0.3990$ &0.6236 $\pm0.3746$ &0.8894 $\pm0.0518$ &\textbf{0.9535} $\pm0.0291$ \\
    \midrule
    ResNet50 + cDCM Loss &\textbf{0.6856} $\pm0.1852$ &0.7991 $\pm0.2693$ &0.7477 $\pm0.2140$ &0.900 $\pm0.0199$ &0.9498 $\pm0.0250$ \\
    \midrule
    Custom Model + cDCM Loss &0.5504 $\pm\textbf{0.1586}$ &\textbf{0.9201} $\pm\textbf{0.0679}$ &\textbf{0.7934} $\pm\textbf{0.0417}$ &0.8460 $\pm0.0885$ &0.9470 $\pm\textbf{0.0226}$ \\
    \bottomrule
\end{tabular}
\end{table*}

For our custom model structure, we also tested a binary cross-entropy loss function and the Deep SAD loss function. Table~\ref{table:some_tests} shows the performance in Fold 1 is poor for the two loss functions, especially the performance on the recall metric. After obtaining two folds of results, we have stopped training more folds because the overall performance will also be poor on average and our main goal is to achieve a good recall.

\begin{table}[hbt]
\caption{Some test results of custom model with BCE and Deep SAD loss.}
  \label{table:some_tests}
\begin{tabular}{lllllll}
    \toprule
    Loss &Fold &Precision &Recall &$F_2$ &Accuracy &AUCROC \\
    \midrule
    BCE &0 &0.7186 &0.7398 &0.7354 &0.9058 &0.9420 \\
    \midrule
    BCE &1 &0.6330 &\textcolor{red}{0.1869} &0.2175 &0.8573 &0.9120 \\
    \midrule
    Deep SAD &0 &0.4474 &0.9049 &0.7512 &0.7924 &0.9198 \\
    \midrule
    Deep SAD &1 &0.7377 &\textcolor{red}{0.4878} &0.5232 &0.8938 &0.9190 \\
    \bottomrule
\end{tabular}
\end{table}

\subsection{Ablation Study}
\label{sec:ablation_study}

We perform an ablation study to investigate dilated convolution, feature fusion, and sequeeze-and-excitation blocks of our custom model structure. 
%
Our ablation study evaluates the model structure on our PPMR dataset, because our custom model structure is too complex and not suitable for our modified CIFAR-10 dataset. We insert model components one by one to illustrate whether the component increases overall performance. Results are shown in Table~\ref{table:ablation_study}. The basic model structure is dilated convolution neural networks, and then we add feature fusion, and finally we add squeeze-and-excitation blocks. We can see that the averaged recall and f-measure of 5-fold CV increases gradually, and the standard deviation of recall and precision decreases gradually. Overall, after adding feature fusion and squeeze-and-excitation blocks, the recall and $F_2$ measure of our model becomes higher and more stable. 
Although precision decreases after adding two new components, the $F_2$ measure increases gradually showing that the model improves overall. Since our PPMR dataset is imbalanced, accuracy is not very meaningful and hence we consider it not in detail. Moreover, AUCROC of the final combined model is also highest amongst the evaluated variants. Hence the ablation study demonstrates the importance of each component in our custom model structure.

\begin{table*}[hbt]
\caption{Model Structure Ablation Study. These models are trained based on our novel cDCM loss function.}
\label{table:ablation_study}
\begin{tabular}{llllll} 
\toprule
Model Blocks &Precision &Recall &$F_2$ measure &Accuracy &AUCROC \\
\midrule
DilatedCNN &\textbf{0.7336} $\pm0.2171$ &0.5612 $\pm0.3166$ &0.5513 $\pm0.2873$ &\textbf{0.8815} $\pm\textbf{0.0442}$ &0.9425 $\pm0.0240$ \\
\midrule
DilatedCNN +FeatureFusion &0.5563 $\pm0.2013$ &0.8747 $\pm0.1325$ &0.7560 $\pm0.0668$ &0.8308 $\pm0.1141$ &0.9347 $\pm\textbf{0.0165}$ \\
\midrule
DilatedCNN +FeatureFusion +SE &0.5504 $\pm\textbf{0.1586}$ &\textbf{0.9201} $\pm\textbf{0.0679}$ &\textbf{0.7934} $\pm\textbf{0.0417}$ &0.8460 $\pm0.0885$ &\textbf{0.9470} $\pm0.0226$ \\
\bottomrule
\end{tabular}
\end{table*}

\section{Conclusions and Future Work}
\label{chapter:conclusion_and_future_work}

Inexperienced readers could misclassify polymicrogyria (PMG) MRIs as normal MRIs in practice,
and a computer-aided method to detect PMG in MRIs is would be of value. 
In the clinical setting, it is important to identify all cases of PMG, so a high recall is important.  
Our model provides a high recall and an acceptable precision and is therefore promising in the future evaluation of PMG cases.

We find cross-entropy-based loss functions predict the majority of images as normal, which causes results of high precision but low recall. In addition, models with cross-entropy-based loss functions easily lead to overfitting and they lack the ability to generalize, especially, if the sample contains unseen features during training. Moreover, one cannot trade-off precision and recall by simply moving the decision threshold in cross-entropy-based loss functions because of out-of-distribution samples. However, our novel cDCM loss function tends to predict images with high $F_2$ measure, which means relatively high recall and acceptable precision, and partly deals with the generalization problem. Finally, our custom model with our novel cDCM loss function can achieve $92.01\%$ recall at $55.04\%$ precision on PPMR dataset.
After comparison, we conclude that our custom model can achieve better performance than EfficientNet and can get more stable results than ResNet50 with our cDCM loss.

This research has a number of shortcomings. Radiologists will find a substantial 
number of normal MRIs in the set of possible PMG MRIs predicted by our algorithm since the precision of our final results is $55.04\%$. Furthermore, because of the limited size of our PPMR dataset, it is challenging to assess how well the model will perform on the actual PMG distribution. 
To our knowledge, this study is the first to distinguish PMG MRIs from normal MRIs by machine learning methods and hence there is a lot of work that needs to be done in the future.

As for future work, our method is promising for some other kinds of medical imaging classification tasks, especially for some tasks where out-of-distribution samples will be encountered when deployed. 
However, precision needs be increased further 
without further decreasing recall.
To achieve this goal, more effective data augmentation and improving the model structure can be explored further.








\section*{Declaration of competing interest}

None Declared.

\section*{Acknowledgment}

Jochen Lang acknowledges the support of the Natural Sciences and Engineering Research Council of Canada (NSERC).

\appendix

\section{Distribution of the PPMR Dataset}\label{appendix_2}

\begin{figure}[htb]
    \centering
    \includegraphics[width=0.95\columnwidth]{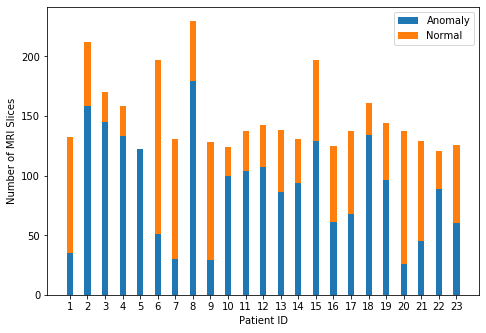}
    \caption{The amount of normal slices and anomaly slices for each patient}
    \label{fig:patients}
\end{figure}

\begin{figure}[htb]
    \centering
    \includegraphics[width=0.95\columnwidth]{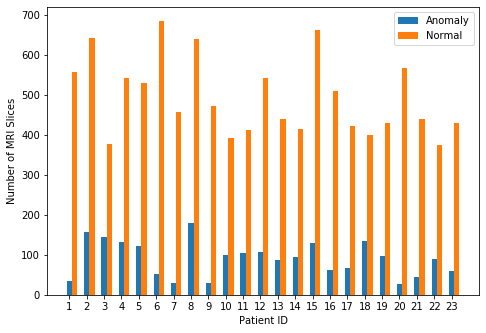}
    \caption{Our hospital polymicrogyria dataset MRI slices distribution}
    \label{fig:patients_with_controls}
\end{figure}

The ratio between the amount of normal MRI slices and the amount of anomaly MRI slices for each patient are shown in Figure~\ref{fig:patients}. And the whole amount of normal MRI slices and anomaly MRI slices for each patient with 3 controls are shown in Figure~\ref{fig:patients_with_controls}. We can see this is an imbalanced dataset.

\section{Choosing the Center and Margin for our cDCM Loss}
\label{sec:DifferentCenter}

One of the key advantages of our proposed cDCM loss is the fact that a center in the latent space can be chosen randomly. Below we give a proof to elucidate why this is correct, followed by an experimental verification. The margin is another important hyper-parameter in our novel cDCM loss function. In Section~\ref{sec:DifferentMarginsComparison}, we experimental show that this margin can also be chosen randomly within reason.

\subsection{Theoretical Considerations}
\label{sec:proof}

\textbf{Proposition:} Given a specific deep learning structure, ideally, if one center $c'$ is initialized randomly and is fixed during training, one model trained on this center $c'$ can achieve the same optimum as other models trained on other centers. It is not necessary to find the true center $c$ first.

\textbf{Proof:}
\begin{enumerate}
\item Assume there is a true center $c(c_0, c_1, c_2, ..., c_n)$ of the latent representations of all normal samples, and $c$ is a n+1 dimensional vector

\item Our randomly defined center is $c'(c_0+s_0, c_1+s_1, c_2+s_2, ..., c_n+s_n)$, where $s_n$ is a constant value.

\item Assume there is an {\em hypothetical} well-trained model $f$ based on the correct center $c$, so $ f(x) = (l_0, l_1, l_2, ..., l_n) $. $l_i$ represents the $i^{th}$ value of the latent representation $l$ in dimension $i$.

\item Our well-trained model based on our randomly initialized center $c'$ can be represented as $ f'(x) = (l_0', l_1', l_2', ..., l_n') $.

\item The squared distance between the latent representation of a sample in the hypothetical well-trained model $f$ from the true center is $ d^2 = (l_0-c_0)^2 + (l_1-c_1)^2 + (l_2-c_2)^2 + ... + (l_n-c_n)^2 $

\item And the squared distance between the latent representation of a sample in our well-trained model $f'$ from the randomly defined center is $ d'^2 = (l_0'-c_0-s_0)^2 + (l_1'-c_1-s_1)^2 + (l_2'-c_2-s_2)^2 + ... + (l_n'-c_n-s_n)^2 $

\item If there exists a simple method to ensure that $ d^2 == d'^2$ for each sample, our well-trained model $f'$ based on center $c'$ will have the same global loss value as the well-trained model $f$ based on the center $c$. An obvious solution would be
\begin{eqnarray} 
l_0 &==& l_0'-s_0 \nonumber\\ 
l_1 &==& l_1'-s_1 \nonumber\\
l_2 &==& l_2'-s_2 \nonumber\\
& &\cdots \nonumber\\
l_n &==& l_n'-s_n \; .\nonumber
\end{eqnarray}

\item We can write $l_n'$ as $l_n + k_n$, where $k_n$ is another constant value. If we want $ l_n == l_n'- s_n $, we must ensure $l_n == l_n' + k_n - s_n$ and hence we have to set $k_n == s_n$.

\item We consider $s_n$ as the shift of the true center $c$. And intuitively, $k_n$ can be regarded as the shift of latent representations in the latent space. A shift of the latent representations can be achieved by the bias term of the last fully-connected layer in the model. Hence, although the loss is only minimized for all training samples as a whole, there exists a simple solution for deep learning models to meet $ d^2 == d'^2$ for each sample by modifying only the bias term.

\item The minimum loss value is 0 given our novel cDCM loss function. In this case, all normal samples will be located at the position of the center in the latent representation.

\item Hence, shifting centers means shifting all normal samples in the same direction and it is not necessary to find the true center based on normal samples.

\end{enumerate}

Overall, the training process could simply ensure that the the correct center $c$ shifts to random center $c'$, and the latent representation of each sample based on the correct center $c$ will also be shifted, to keep the same distance from each data sample to the center. However, if the loss does not reach 0 for the normal samples, and considering the loss caused by the anomaly samples, there could be other complex transformations for these latent representations to make sure the global loss is optimal.

In addition, different model structures likely map samples differently into the latent space and hence some may be better than others. Therefore, model structure is also important to achieve good performance.

\subsection{Experimental Comparison of Different Centers}
\label{sec:DifferentCentersComparison}

On modified CIFAR-10 dataset, we try three different centers during our novel cDCM loss model training. The first center 128-dimensional all zero vector; the second center is all ones, and the last one is random value drawn from an uniform sampling distribution in the range between 0.0 to 1.0. In Appendix~\ref{sec:DifferentCentersComparison}, we show that the center of our novel cDCM loss can be chosen randomly under the assumption that the model is well trained. In our experiments, we also show that given a specific model structure, choosing different centers in initialization can achieve almost the same results. In Table~\ref{table:different_center_comparison_results_table}, we can see performance results, especially AUCROC, for different centers are almost the same in all classes. The small difference are likely caused by weight initialization and shuffling batches during training. We save the model with the best validation AUCROC value but a small AUCROC value difference may move the decision threshold by a relatively large amount. This is the likely cause for the performance metrics, except for AUCROC, exhibit small differences in some cases with different centers. 
The overall results clearly demonstrate that the center can be chosen randomly in practice.

\begin{table*}[hbt]
\caption{Comparison between different centers of our novel cDCM loss function on the modified CIFAR-10 dataset. Each result value in the table is the average of 5 experiments with different weight initialization.}
  \label{table:different_center_comparison_results_table} 
  \begin{tabular}{lllllll}
    \toprule
    Normal Class & Center & Precision & Recall & $F_2$ measure & Accuracy & AUCROC \\
    \midrule
    airplane & All 0. & 0.9837 $\pm0.0044$ & 0.7264 $\pm0.0456$ & 0.7661 $\pm0.0400$ & 0.7427 $\pm0.0377$ & 0.8896 $\pm0.0088$ \\
    airplane & All 1. & 0.9841 $\pm0.0048$ & 0.7239 $\pm0.0594$ & 0.7637 $\pm0.0530$ & 0.7407 $\pm0.0496$ & 0.8929 $\pm0.0058$ \\
    airplane & Random & 0.9856 $\pm0.0024$ & 0.7147 $\pm0.0357$ & 0.7560 $\pm0.0318$ & 0.7338 $\pm0.0304$ & 0.8948 $\pm0.0040$ \\
    \midrule
    automobile & All 0. & 0.9918 $\pm0.0031$ & 0.6888 $\pm0.0318$ & 0.7334 $\pm0.0285$ & 0.7147 $\pm0.0265$ & 0.8983 $\pm0.0050$ \\
    automobile & All 1. & 0.9934 $\pm0.0011$ & 0.6735 $\pm0.0172$ & 0.7198 $\pm0.0157$ & 0.7021 $\pm0.0151$ & 0.8927 $\pm0.0050$ \\
    automobile & Random & 0.9946 $\pm0.0015$ & 0.6568 $\pm0.0397$ & 0.7044 $\pm0.0364$ & 0.6879 $\pm0.0347$ & 0.8946 $\pm0.0050$ \\
    \midrule
    bird & All 0. & 0.9530 $\pm0.0074$ & 0.6004 $\pm0.0289$ & 0.6482 $\pm0.0267$ & 0.6136 $\pm0.0227$ & 0.7225 $\pm0.0247$ \\
    bird & All 1. & 0.9576 $\pm0.0091$ & 0.5641 $\pm0.0552$ & 0.6140 $\pm0.0518$ & 0.5849 $\pm0.0436$ & 0.7310 $\pm0.0190$ \\
    bird & Random & 0.9548 $\pm0.0037$ & 0.5754 $\pm0.0480$ & 0.6246 $\pm0.0449$ & 0.5932 $\pm0.0398$ & 0.7261 $\pm0.0138$ \\
    \midrule
    cat & All 0. & 0.9703 $\pm0.0043$ & 0.5412 $\pm0.0495$ & 0.5932 $\pm0.0470$ & 0.5720 $\pm0.0414$ & 0.7725 $\pm0.0126$ \\
    cat & All 1. & 0.9662 $\pm0.0038$ & 0.6053 $\pm0.0479$ & 0.6538 $\pm0.0448$ & 0.6256 $\pm0.0396$ & 0.7741 $\pm0.0056$ \\
    cat & Random & 0.9745 $\pm0.0020$ & 0.5273 $\pm0.0376$ & 0.5803 $\pm0.0362$ & 0.5621 $\pm0.0323$ & 0.7786 $\pm0.0145$ \\
    \midrule
    deer & All 0. & 0.9711 $\pm0.0068$ & 0.6908 $\pm0.0627$ & 0.7324 $\pm0.0558$ & 0.7029 $\pm0.0504$ & 0.8378 $\pm0.0062$ \\
    deer & All 1. & 0.9728 $\pm0.0018$ & 0.6546 $\pm0.0418$ & 0.7002 $\pm0.0385$ & 0.6727 $\pm0.0358$ & 0.8261 $\pm0.0058$ \\
    deer & Random & 0.9750 $\pm0.0033$ & 0.6275 $\pm0.0374$ & 0.6754 $\pm0.0345$ & 0.6502 $\pm0.0311$ & 0.8310 $\pm0.0074$ \\
    \midrule
    dog & All 0. & 0.9821 $\pm0.0020$ & 0.6717 $\pm0.0213$ & 0.7169 $\pm0.0192$ & 0.6934 $\pm0.0177$ & 0.8688 $\pm0.0009$ \\
    dog & All 1. & 0.9852 $\pm0.0033$ & 0.6231 $\pm0.0402$ & 0.6722 $\pm0.0372$ & 0.6522 $\pm0.0337$ & 0.8624 $\pm0.0043$ \\
    dog & Random & 0.9840 $\pm0.0030$ & 0.6415 $\pm0.0490$ & 0.6891 $\pm0.0451$ & 0.6678 $\pm0.0417$ & 0.8645 $\pm0.0032$ \\
    \midrule
    frog & All 0. & 0.9880 $\pm0.0067$ & 0.7341 $\pm0.0384$ & 0.7736 $\pm0.0335$ & 0.7526 $\pm0.0307$ & 0.9124 $\pm0.0186$ \\
    frog & All 1. & 0.9909 $\pm0.0016$ & 0.7090 $\pm0.0142$ & 0.7518 $\pm0.0126$ & 0.7323 $\pm0.0117$ & 0.9107 $\pm0.0056$ \\
    frog & Random & 0.9887 $\pm0.0048$ & 0.7477 $\pm0.0571$ & 0.7855 $\pm0.0502$ & 0.7651 $\pm0.0476$ & 0.9156 $\pm0.0036$ \\
    \midrule
    horse & All 0. & 0.9863 $\pm0.0020$ & 0.6030 $\pm0.0104$ & 0.6538 $\pm0.0098$ & 0.6352 $\pm0.0097$ & 0.8604 $\pm0.0144$ \\
    horse & All 1. & 0.9847 $\pm0.0058$ & 0.6120 $\pm0.0689$ & 0.6613 $\pm0.0647$ & 0.6420 $\pm0.0583$ & 0.8510 $\pm0.0172$ \\
    horse & Random & 0.9868 $\pm0.0036$ & 0.5856 $\pm0.0567$ & 0.6368 $\pm0.0536$ & 0.6198 $\pm0.0485$ & 0.8489 $\pm0.0176$ \\
    \midrule
    ship & All 0. & 0.9836 $\pm0.0040$ & 0.8684 $\pm0.0263$ & 0.8891 $\pm0.0214$ & 0.8685 $\pm0.0203$ & 0.9395 $\pm0.0061$ \\
    ship & All 1. & 0.9880 $\pm0.0028$ & 0.8281 $\pm0.0331$ & 0.8557 $\pm0.0280$ & 0.8362 $\pm0.0279$ & 0.9412 $\pm0.0062$ \\
    ship & Random & 0.9859 $\pm0.0030$ & 0.8618 $\pm0.0303$ & 0.8839 $\pm0.0250$ & 0.8645 $\pm0.0245$ & 0.9439 $\pm0.0042$\\
    \midrule
    truck & All 0. & 0.9865 $\pm0.0015$ & 0.7231 $\pm0.0214$ & 0.7638 $\pm0.0191$ & 0.7419 $\pm0.0193$ & 0.9019 $\pm0.0085$ \\
    truck & All 1. & 0.9828 $\pm0.0020$ & 0.7610 $\pm0.0298$ & 0.7968 $\pm0.0260$ & 0.7729 $\pm0.0250$ & 0.8974 $\pm0.0076$ \\
    truck & Random & 0.9830 $\pm0.0032$ & 0.7783 $\pm0.0325$ & 0.8119 $\pm0.0281$ & 0.7882 $\pm0.0266$ & 0.9063 $\pm0.0063$ \\
    \bottomrule
\end{tabular}
\end{table*}

\subsection{Experimental Comparison of Different Margins}
\label{sec:DifferentMarginsComparison}

On the modified CIFAR-10 dataset, we try two different margins during our novel cDCM loss model training, which are 5 and 10. In Table~\ref{table:different_margin_comparison_results_table}, AUCROC values in these two margins are almost same. In addition, we can see in some normal classes, like automobile, deer, and etc., the $F_2$ measure on margin 5 is slightly better than the $F_2$ measure on margin 10, while in some other normal classes, like dog, frog, and etc., the $F_2$ measure on margin 10 is slightly better than the $F_2$ measure on margin 5. There is no clear evidence to show which margin is the best, but choosing different margins would not affect the final results too much. Based on our novel cDCM loss function, setting the margin too small results in a tiny hypersphere, while setting the margin too big results in a huge hypersphere. In the worst scenario, setting margin to 0 or infinity is meaningless. Hence, hyper-parameter "margin" should be set in a reasonable manner.

\begin{table*}[hbt]
\caption{Comparison between different margins of our novel cDCM loss function on the modified CIFAR-10 dataset. We use the center of 'All 1.' here. Each result value in the table is the average of 5 experiments with different weight initialization.}
  \label{table:different_margin_comparison_results_table} 
\begin{tabular}{lllllll}
    \toprule
    Normal Class & Margin & Precision & Recall & $F_2$ measure & Accuracy & AUCROC \\
    \midrule
    airplane & 5 & 0.9841 $\pm0.0048$ & 0.7239 $\pm0.0594$ & 0.7637 $\pm0.0530$ & 0.7407 $\pm0.0496$ & 0.8929 $\pm0.0058$ \\
    airplane & 10 & 0.9823 $\pm0.0030$ & 0.7328 $\pm0.0438$ & 0.7717 $\pm0.0391$ & 0.7476 $\pm0.0374$ & 0.8891 $\pm0.0093$ \\
    \midrule
    automobile & 5 & 0.9934 $\pm0.0011$ & 0.6735 $\pm0.0172$ & 0.7198 $\pm0.0157$ & 0.7021 $\pm0.0151$ & 0.8927 $\pm0.0050$ \\
    automobile & 10 & 0.9943 $\pm0.0017$ & 0.6500 $\pm0.0197$ & 0.6983 $\pm0.0180$ & 0.6816 $\pm0.0167$ & 0.8951 $\pm0.0078$ \\
    \midrule
    bird & 5 & 0.9576 $\pm0.0091$ & 0.5641 $\pm0.0552$ & 0.6140 $\pm0.0518$ & 0.5849 $\pm0.0436$ & 0.7310 $\pm0.0190$ \\
    bird & 10 & 0.9549 $\pm0.0049$ & 0.5776 $\pm0.0710$ & 0.6263 $\pm0.0653$ & 0.5951 $\pm0.0582$ & 0.7323 $\pm0.0132$ \\
    \midrule
    cat & 5 & 0.9662 $\pm0.0038$ & 0.6053 $\pm0.0479$ & 0.6538 $\pm0.0448$ & 0.6256 $\pm0.0396$ & 0.7741 $\pm0.0056$ \\
    cat & 10 & 0.9623 $\pm0.0089$ & 0.6230 $\pm0.0595$ & 0.6695 $\pm0.0540$ & 0.6382 $\pm0.0460$ & 0.7730 $\pm0.0114$ \\
    \midrule
    deer & 5 & 0.9728 $\pm0.0018$ & 0.6546 $\pm0.0418$ & 0.7002 $\pm0.0385$ & 0.6727 $\pm0.0358$ & 0.8261 $\pm0.0058$ \\
    deer & 10 & 0.9754 $\pm0.0032$ & 0.6280 $\pm0.0452$ & 0.6758 $\pm0.0411$ & 0.6508 $\pm0.0378$ & 0.8308 $\pm0.0107$ \\
    \midrule
    dog & 5 & 0.9852 $\pm0.0033$ & 0.6231 $\pm0.0402$ & 0.6722 $\pm0.0372$ & 0.6522 $\pm0.0337$ & 0.8624 $\pm0.0043$ \\
    dog & 10 & 0.9828 $\pm0.0004$ & 0.6580 $\pm0.0118$ & 0.7046 $\pm0.0108$ & 0.6818 $\pm0.0105$ & 0.8652 $\pm0.0051$ \\
    \midrule
    frog & 5 & 0.9909 $\pm0.0016$ & 0.7090 $\pm0.0142$ & 0.7518 $\pm0.0126$ & 0.7323 $\pm0.0117$ & 0.9107 $\pm0.0056$ \\
    frog & 10 & 0.9867 $\pm0.0017$ & 0.7602 $\pm0.0222$ & 0.7967 $\pm0.0195$ & 0.7750 $\pm0.1920$ & 0.9123 $\pm0.0073$ \\
    \midrule
    horse & 5 & 0.9847 $\pm0.0058$ & 0.6120 $\pm0.0689$ & 0.6613 $\pm0.0647$ & 0.6420 $\pm0.0583$ & 0.8510 $\pm0.0172$ \\
    horse & 10 & 0.9884 $\pm0.0058$ & 0.5724 $\pm0.0549$ & 0.6245 $\pm0.0516$ & 0.6089 $\pm0.0459$ & 0.8606 $\pm0.0109$ \\
    \midrule
    ship & 5 & 0.9880 $\pm0.0028$ & 0.8281 $\pm0.0331$ & 0.8557 $\pm0.0280$ & 0.8362 $\pm0.0279$ & 0.9412 $\pm0.0062$ \\
    ship & 10 & 0.9870 $\pm0.0055$ & 0.8448 $\pm0.0460$ & 0.8695 $\pm0.0378$ & 0.8501 $\pm0.0364$ & 0.9465 $\pm0.0026$ \\
    \midrule
    truck & 5 & 0.9828 $\pm0.0020$ & 0.7610 $\pm0.0298$ & 0.7968 $\pm0.0260$ & 0.7729 $\pm0.0250$ & 0.8974 $\pm0.0076$ \\
    truck & 10 & 0.9785 $\pm0.0041$ & 0.8023 $\pm0.0297$ & 0.8321 $\pm0.0251$ & 0.8061 $\pm0.0241$ & 0.9030 $\pm0.0110$ \\
    \bottomrule
\end{tabular}
\end{table*}

\section{Tests of Statistical Significance}
\label{sec:tests_of_statistical_significance}

\begin{figure}[htb]
    \centering
    \includegraphics[width=0.95\columnwidth]{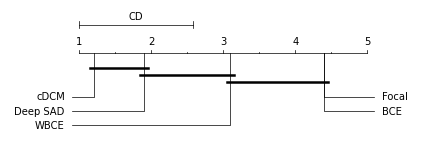}
    \caption{Critical difference diagram for the pair-wise Bonferroni-Dunn Test}
    \label{fig:bonferroni_dunn_test}
\end{figure}

To verify whether our cDCM loss function has a significant difference from other loss functions on the modified CIFAR-10 dataset, we conduct a Friedman Test~\cite{friedman1940comparison} of statistical significance on the $F_2$ measure metric and the Bonferroni-Dunn Test for pair-wise statistical significance. Based on the Friedman test, we can reject the null hypothesis that there is no significance difference between the loss function at the 0.05 threshold (p-value is $9.34e^{-7}$). The critical difference diagram for the pair-wise Bonferroni-Dunn Test in shown in Figure~\ref{fig:bonferroni_dunn_test}. We can see the cDCM loss function is significantly better than BCE, WBCE and Focal loss functions, but the test shows no significant difference to the Deep SAD loss function at $\alpha=0.05$.

\bibliographystyle{unsrt}  
\bibliography{references.bib}

\end{document}